\documentclass[runningheads]{llncs}
\usepackage{amsmath}

\usepackage{wrapfig, booktabs}
\usepackage{graphicx}
\usepackage{subfigure}
\usepackage{amsmath,amssymb}
\usepackage{times}
\usepackage{epsfig}
\usepackage{graphicx}
\usepackage{amsmath}
\usepackage{amssymb}
\usepackage{dirtytalk}
\usepackage{xcolor}
\usepackage{cite}
\usepackage{graphicx}
\usepackage{subfigure}

\usepackage{amsmath}

\begin{document}
\title{DeepSSM: A Blueprint for Image-to-Shape Deep Learning Models}
%

%
\author{Riddhish Bhalodia  \and Shireen Elhabian \and Jadie Adams \and Wenzheng Tao \and Ladislav Kavan \and Ross Whitaker}

     
%
\institute{School of Computing, University of Utah, UT, USA }

\maketitle              
\begin{abstract}
Statistical shape modeling (SSM) characterizes anatomical variations in a population of shapes generated from medical images. Statistical analysis of shapes requires consistent shape representation across samples in shape cohort. Establishing this representation entails a processing pipeline that includes anatomy segmentation, image re-sampling, shape-based registration, and non-linear, iterative optimization. These shape representations are then used to extract low-dimensional shape descriptors that are anatomically relevant to facilitate subsequent statistical analyses in different applications. However, the current process of obtaining these shape descriptors from imaging data relies on human and computational resources, requiring domain expertise for segmenting anatomies of interest. Moreover, this same taxing pipeline needs to be repeated to infer shape descriptors for new image data using a pre-trained/existing shape model. Here, we propose DeepSSM, a deep learning-based framework for learning the functional mapping from images to low-dimensional shape descriptors and their associated shape representations, thereby inferring statistical representation of anatomy directly from 3D images. Once trained using an existing shape model, DeepSSM circumvents the heavy and manual pre-processing and segmentation required by classical models and significantly improves the computational time, making it a viable solution for fully end-to-end shape modeling applications. In addition, we introduce a model-based data-augmentation strategy to address data scarcity, a typical scenario in shape modeling applications. Finally, this paper presents and analyzes two different architectural variants of DeepSSM with different loss functions using three medical datasets and their downstream clinical application. Experiments showcase that DeepSSM performs comparably or better to the state-of-the-art SSM both quantitatively and on application-driven downstream tasks. Therefore, DeepSSM aims to provide a comprehensive blueprint for deep learning-based image-to-shape models.
\end{abstract}
\section{Introduction}
\label{sec:intro}

Statistical shape modeling (SSM) enables the quantitative study of anatomical forms in medical and biological sciences. Shape models provide a geometrical description of each shape that is statistically consistent across the studied shape population. A classical approach for SSM is to use landmark points to represent shapes, where landmark points often correspond to specific anatomical or morphological features. Landmark-based SSM was first described in the pioneering work of D'Arcy Thomson \cite{thompson1942growth}. These landmark correspondences provide a flexible and easy-to-use shape representation to capture population variability using statistical methods such as principal component analysis (PCA) to derive shape descriptors that capture anatomically relevant morphological features. More recent works have shown a push towards obtaining dense sets of correspondences across the population automatically \cite{cates2007shape, davies2002MDL, styner2006spharm}. Another method of quantifying shape differences is using the coordinate transformations that align the population of images/shapes to a predefined atlas \cite{beg2005computing}. These automated methods for representing shapes and their subsequent analysis finds application in several medical domains, such as orthopedics \cite{harris2013cam, bhalodia2020quantifying}, implant design \cite{goparaju2018evaluation, kozic2010optimisation}, neuroscience \cite{greig2001brain, zhao2008hippocampus}, and cardiology \cite{cates2013afib, bhalodia2018endtoend}. 

The current pipeline for shape analysis relies on obtaining a shape representation (e.g., correspondences, transformations) and the associated shape descriptors, followed by an application-dependent use of these shape descriptors. This \emph{shape descriptor} is typically a low-dimensional representation learned from the population shape data; PCA scores is an example of such shape descriptor. Subsequent analyses are application-dependent and may vary from binary prediction of a pathology \cite{gutierrez2018deep}, a localized detection of a morphological defect \cite{harris2013cam}, or understanding subtleties of the morphology and testing the associated scientific hypotheses \cite{cates2018afib}. 

 \emph{Point distribution models} (PDMs) \cite{RTW:Gre91} relies on dense correspondences for shape representation, which are geometrically consistent set of particles placed on each anatomical surface in the shape population. The position of these correspondences can be optimized across the population via using different metrics, such as entropy \cite{cates2007shape} or minimum description length \cite{davies2002MDL}. These metrics aim to reduce the statistical complexity of the model and ensure a compact statistical representation. SPHARM-PDM \cite{styner2006spharm}, on the other hand, utilizes a parametric representation of the surface using spherical harmonics. These state-of-the-art correspondence models require anatomy segmentation followed by a lengthy pipeline of pre-processing steps such as image re-sampling, rigid alignment, and smoothing, followed by a computationally expensive optimization to establish shape correspondences. These time- and resources-intensive steps are not only necessary for learning the shape model, but also for inferring shape descriptors for a new image given a pre-trained shape model, rendering such models ineffective to be used in a fully automated pipeline. 

On the other hand, deformation fields can represent shapes directly from images via image-based registration, and hence alleviate the need for anatomy segmentation. However, the shape representation via implicit transformations between images is not intuitive and can be challenging for clinicians to understand and interpret. Additionally, to obtain a concrete shape descriptor for a single shape, we need a notion of an atlas image, which is not always easily obtained. Furthermore, to accurately represent only the shape information, we need to have anatomy segmentation, or we need a way to disentangle shape and background components from the deformation fields. Due to these drawbacks, especially the lack of easily interpretable shape representation, this paper, focuses only on PDM based shape representation models.


This paper addresses the problems in existing PDM models by proposing a novel framework, DeepSSM. The framework proposes deep convolutional neural network (CNN) architectures that learn to discover the low-dimensional shape descriptor and the associated correspondences directly from images. DeepSSM allows for a fully end-to-end inference pipeline that bypasses the intensive pre-processing and segmentation that requires manual supervision. Additionally, the model's inference is computationally fast compared to the heavy optimization in state-of-the-art PDM approaches. To train the deep network robustly with limited data ---a typical challenge in medical imaging applications--- we also introduce a model-based data-augmentation strategy that enables the generation of thousands of training samples that preserve population-specific statistics. Furthermore, this paper provides a comprehensive and systemic approach for developing supervised deep learning-based models for image-to-shape inference tasks using two different architectural variants along with loss functions and strategies tailored for individual datasets and subsequent clinical applications. 

\begin{figure}[!h]
    \centering
    \includegraphics[width=0.9\textwidth]{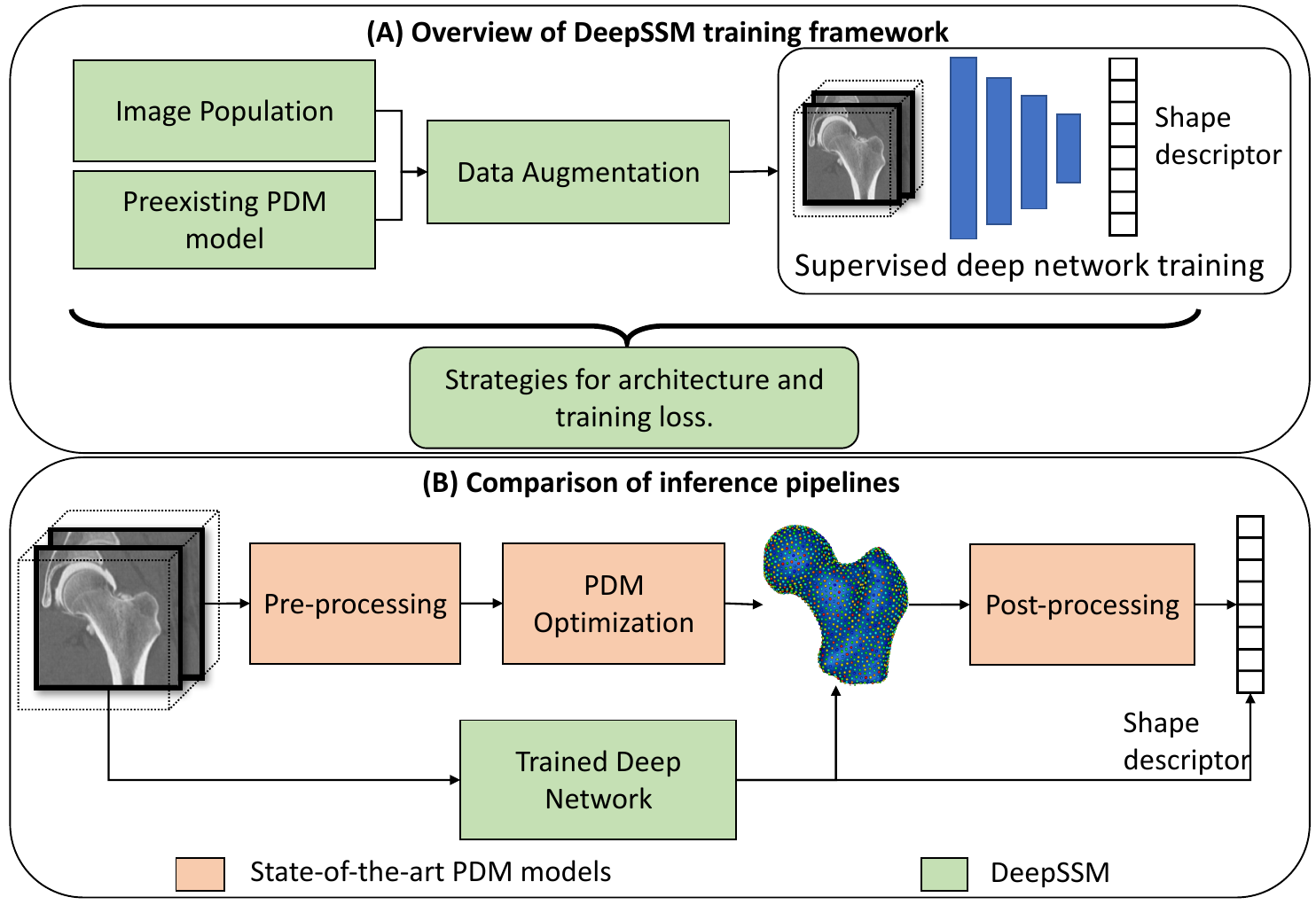}
    \caption{\textbf{DeepSSM:} (A) shows the overall framework presented in this paper, which includes data-augmentation and deep network architectures and training strategies to handle wide range of shape data and clinical applications. (B) shows the comparison between inference pipelines of a state-of-the-art PDM and DeepSSM.}
    \label{fig:teaser}
\end{figure}

A preliminary variant of this work has been presented in conference papers \cite{bhalodia2018endtoend} and \cite{bhalodia2018deepssm}. Here, we significantly extend and complement this framework by proposing the following novel additions:
\begin{itemize}
    \item A modified, more general approach for data-augmentation.
    \item Proposed a novel architectural variant for obtaining a non-linear shape descriptor for complex datasets. Modification to the original training loss function, and a novel alternative loss function utilized for certain datasets.
    \item Comprehensive experimentation on three different datasets with improved evaluations - both quantitatively and via downstream usage of the shape descriptors. This experimental spectrum also provides a detailed analysis of which architecture/loss function to be used for each dataset or application.
\end{itemize}

In summary, DeepSSM provides a blueprint for inferring population-specific statistical representation of shapes from images. This blueprint proposes data augmentation to handle limited data, different neural network architectures for various applications, and training strategies to learn shape descriptors. Finally, it includes a set of examples and guidelines for using this proposed architecture in real-world SSM applications, providing a launching point for all further image-to-shape models. Figure \ref{fig:teaser} illustrates DeepSSM as a framework and its inference pipeline compared to the standard correspondence-based method.
\section{Related Work}
\label{sec:literature}



The DeepSSM framework learns the function that maps input images onto a landmark/correspondence-based shape space. In related literature, explicit correspondences have been expressed using geometric parametrizations \cite{davies2002MDL, RTW:Sty2000} as well as functional maps \cite{ovsjanikov2012functional}. Furthermore, particle distribution models \cite{cates2007shape, davies2002MDL, styner2006spharm} are cumbersome and require manual input. DeepSSM learns a PDM model and can be supervised with any given choice of PDM and here we choose to work optimization-based PDM \emph{ShapeWorks} \cite{cates2017shapeworks} since they observe the entire population and lead to less biased models compared to pairwise approaches such as SPHARM \cite{styner2006spharm}. Extensive comparison between such PDMs is presented in \cite{goparaju2018evaluation} which clearly puts \emph{ShapeWorks} as the state-of-the-art PDM model.

Research on atlas construction and image registration is also relevant, these techniques typically involve the use of different parametrization for image-to-image transformations, such as b-splines \cite{rueckert1999nonrigid}, deformable fields with elastic penalty \cite{bajcsy1989multiresolution}, and diffeomorphic static or dynamic flow fields \cite{beg2005computing}. These deformation models have proven useful for shape analysis \cite{joshi2012diffeomorphic, durrleman2014morphometry}, but their use has mostly been exploited for image alignment \cite{beg2005computing}. In recent years, there has been a lot of interest in deep learning based unsupervised registration models \cite{balakrishnan2018unsupervised, yang2017quicksilver} that provide a computationally fast alternative to classical models while maintaining registration accuracy. While these models are applicable directly on images, to successfully utilize them for shape analysis they need to be applied on the segmentation of the anatomy \cite{bhalodia2019cooperative}. Additionally, expressing the shape representation for a single sample requires a notion of an atlas image that might not be always available.

 Convolution neural networks (CNNs) \cite{lecun1998cnn} for 3D volumes has been explored extensively in medical imaging community \cite{wang2019deeply, ronneberger2015unet}. Deep learning has paved  way for several application in detection \cite{zheng2015detection}, classification \cite{li2014classification}, and segmentation \cite{badrinarayanan2015segnet2, ronneberger2015unet}. There are other works that incorporate shape priors such as \cite{ACNN} regularizes anatomy segmentation by learning low-dimensional nonlinear shape representations as a constraint. While \cite{dalca2018anatomical} utilizes a variational autoencoder as a prior generative model on brain label maps for unsupervised segmentation. Regression from 3D volumes, which is a more relevant application for DeepSSM, has been used in limited capacity. In \cite{huang2017heartnet} they regress the orientation and position of the heart from 2D ultrasound, and \cite{cao2018deformable} utilizes regression network to find a displacement vector between nearby patches in order to improve registration model. Another related recent work \cite{milletari2017stats} demonstrates the efficacy of using PCA scores in regression of the landmark positions used for 2D ultrasound segmentation task. Similarly, \cite{Tothova2018} use PCA scores as a prior for probabilistic 2D surface reconstruction and \cite{Tothova2020} extend this to probabilistic 3D surface reconstruction from sparse 2D images. The proposed framework of DeepSSM in this work and the proof of concept described in \cite{bhalodia2018endtoend, bhalodia2018deepssm} extends the idea of using PCA scores for 3D images, as well as explores the construction and use of a non-linear low dimensional shape descriptor to replace PCA scores. DeepSSM also proposed a novel data-augmentation scheme to counter the problem of data availability in medical imaging applications, and provides strategies for using the model under different scenarios involving data or downstream applications.

\section{Methods}
\label{sec:methods}

This section focuses on the proposed DeepSSM framework that entail the data-augmentation and the different DeepSSM variants. The specific details of network architectures and the associated implementation can be found in \ref{app:netdet}.

\subsection{Data Augmentation}
\label{sub:dataaug}

Deep learning models are data-hungry, and data is usually scarce in medical imaging applications; therefore, data-augmentation methods are typically used to significantly increase the training sample size. For statistical shape modeling, the given population size is typically in the range of 100-200 samples; this is insufficient to train a CNN as the network will overfit. To handle this data scarcity issue, DeepSSM frameworks include a model-based data augmentation approach to generate augmented training samples that follow the statistics of the given population.

We start with a set of $N$ \emph{original images} $\mathcal{I} = \{I_1, ..., I_N\}$ and obtain a point distribution model (PDM), also called a correspondence model, on these shapes. We use \emph{ShapeWorks} \cite{cates2017shapeworks} to generate the PDMs used in the experiments presented in this work. In a recent benchmarking study \cite{goparaju2018evaluation,goparaju2020benchmarking}, \emph{ShapeWorks} was shown to discover clinically relevant shape differences. However, any set of correspondences (i.e., a PDM or even manual landmarks) can be utilized for the proposed model. This PDM will give us a set of surface correspondence points for the anatomy associated with each image. Let's denote this set of original correspondence points $\mathcal{ C } = \{ C_1, ..., C_N\}$.

The first step in data augmentation is to define a generative model over the existing PDM space (also known as shape space). This allows us to sample plausible shapes from the distribution. In previous iterations of DeepSSM, we performed PCA on the shape space and modeled it parametrically using a Gaussian model \cite{bhalodia2018deepssm} or a Gaussian mixture model (GMM) \cite{bhalodia2018endtoend}. However, when the underlying distribution of the original shape space is nonlinear, the Gaussian model will ignore vital subtleties in the distribution. Also, the GMM model with too few clusters is prone to overestimating the variance and sampling implausible shapes. 

\begin{figure}
    \centering
    \includegraphics[width=\textwidth]{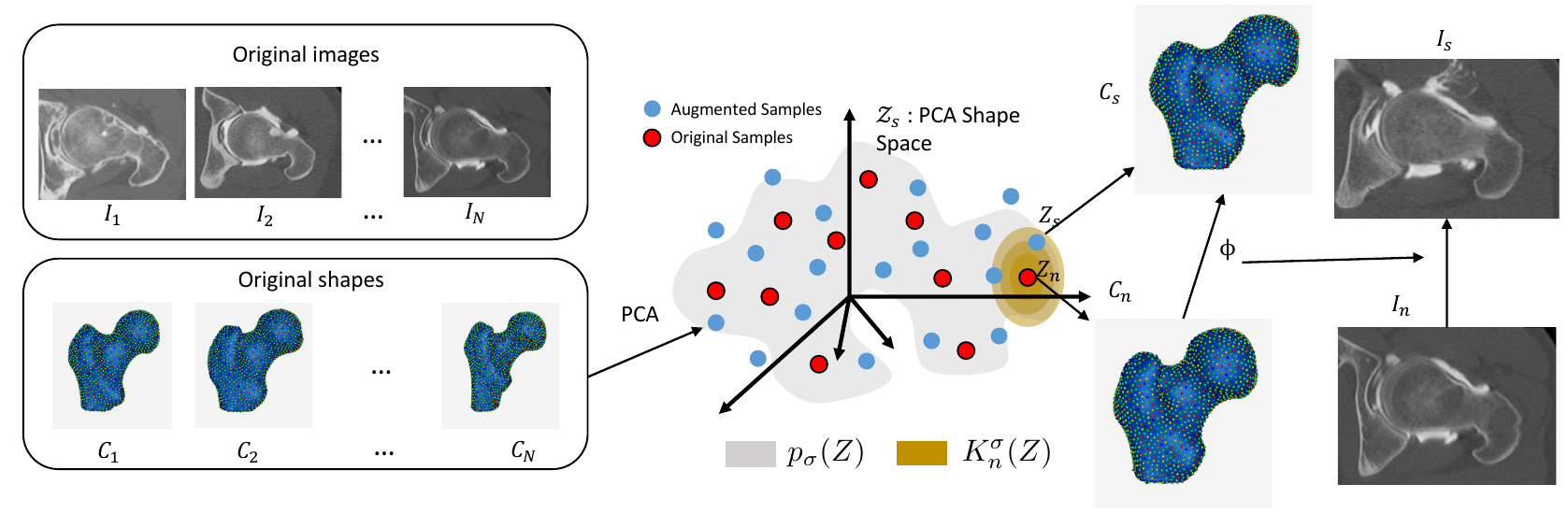}
    \caption{\textbf{Data augmentation pipeline}}
    \label{fig:augmentationShape}
\end{figure}

To counteract these problems, we propose a non-parametric kernel density estimate (KDE) for modeling the PCA space of the correspondences. We found this to be a robust augmentation strategy that generates plausible shapes while still capturing all the subtleties of the original shape space. Augmentation is performed in PCA space and not directly on the shape space as it preserves the smoothness of the anatomy making it a more plausible shape without noise. Additionally, the augmentation is performed by utilizing all available PCA modes, i.e. N-1, which along with proving smoothing will also ensure that we do not destroy any inherent characteristic that shape distribution may exhibit. Let $\mathcal{ Z } = \{ Z_1, ..., Z_N\}$ represent the PCA scores for the correspondences, then we define $p_\sigma(Z)$ as:

\begin{equation}\label{eqn:pz}
  p_\sigma(Z) =\frac{1}{N} \sum_{n=1}^N K_n^\sigma(Z), ~~\operatorname{s.t.}~~K_n^\sigma(Z) = \frac{1}{(2 \pi \sigma^2)^{L/2}}  \exp \left( - \frac{||Z - Z_n||^2} {2\sigma^2}\right),
\end{equation}

where $L$ is the dimension of the PCA subspace. Here the kernel bandwidth, $\sigma$, is computed as the average nearest neighbor distance in the PCA subspace. In this case, when a new PCA score $Z_s \sim p_\sigma(Z)$ is sampled from the kernel of the $n^{th}$ training sample ($K_n^\sigma(Z)$), we can obtain the sampled correspondence, $C_s$, via PCA reconstruction. Once we have the sample correspondence representation, we need the associated image $I_s$ to complete the data augmentation process. We find the original correspondence, $C_n$, from whose kernel $C_s$ is sampled from and use the corresponding raw image $I_n$. Using the Thin plate spline (TPS) warp between $C_n \rightarrow C_s$ (serving as accurate landmarks on the anatomy surface), we warp the image $I_n \rightarrow I_s$. Using this technique, we can generate thousands of training data pairs of images and their corresponding PCA score vectors/correspondence points, while preserving the shape population statistics. The entire pipeline is depicted in Figure \ref{fig:augmentationShape}.

\subsection{DeepSSM Variants} 
\label{sub:netarch}

Using the above data augmentation techniques, we form the training data composed of (i) raw 3D images, (ii) correspondences associated with each scan, and (iii) the PCA scores associated with the correspondences. In the following section, we present the DeepSSM architecture and associated loss functions. We also introduce a novel alternate architecture that utilizes a non-linear representation of correspondences in place of the PCA scores. 

\subsubsection{Basic Framework}

The DeepSSM framework predicts a PCA score representation of the correspondence points from their associated 3D volumes/images. Each input is the image and correspondence pair given by $(I_i, C_i)$, and $Z_i$ is the PCA score of correspondence $C_i$. We use a CNN-based architecture with five convolution layers and two fully connected layers to perform a regression task from 3D raw CT/MRI volume to correspondences (see Figure \ref{fig:netarch} [A]). Detailed architectural description is provided in \ref{app:netdet}. Throughout the paper, we refer to this architecture as \emph{Base-DeepSSM}.
This CNN block is the score encoder given as $f_\theta$ that learns the PCA score shape descriptor $Z_i$. We recover the correspondences from the PCA scores by passing them through a final fully connected layer. This final layer does not have a non-linear activation and is given as $r_\phi = Wz + b$. The weight and bias are kept \emph{fixed} (they are not updated) during the training and are set to the PCA basis and mean shape, respectively. This pair of set weights and bias performs the PCA reconstruction and we obtain the predicted correspondences ($\hat{C}$) as the output as $r_\phi(f_\theta(I)) = \hat{C}$. Therefore, we obtain both the correspondence shape representation as well as the associated shape descriptor (in the form of PCA scores) from the image. Theoretically, the parameters of the last layer can be updated along with the network; however, PCA score as a shape descriptor is more interpretable and is useful in further visualization of modes of shape variation. 

For the training loss, we can directly apply supervision on the correspondences as the network architecture itself handles the PCA encoding. This is given as:
\begin{align}
    \mathcal{L}_{corr} &= \frac{1}{M}\sum\limits_{i=1}^M ||C_i - r_\phi(f_\theta(I_i))||^2
    \label{eq:corr-l2}
\end{align}

\textbf{Note about loss functions: } Initial iterations of DeepSSM provided in \cite{bhalodia2018deepssm, bhalodia2018endtoend}, utilized supervision on the PCA scores directly and defined the loss function as follows:
\begin{align}
 \mathcal{L}_{pca} = \frac{1}{M}\sum\limits_{i=1}^M||Z_i - f_\theta(I_i)||^2 
 \label{eq:pca-l2}
\end{align}

 However, there is a key drawback to learning the PCA scores directly as they are inherently ordered. In the typical setting, the PCA scores will have an arbitrary scale per component. Components that express major modes of variation exhibit numerically larger PCA scores than those that express small and subtle shape variations. Given the nature of neural networks and the loss function given in  Eq. \ref{eq:pca-l2}, the network tends to generalize better on the larger PCA modes than smaller ones, missing subtle shape statistics that might be important for subsequent analysis. The correspondence-based loss function (Eq. \ref{eq:corr-l2}) solves this problem of non-uniform error across the PCA scores, as the correspondence points are equally weighted, so all the shape variations are equally captured. The correspondence-based loss also relies on the PCA shape descriptor; the only difference is that the PCA encoding is taken care of implicitly by the network architecture. In all the datasets presented in Section \ref{sec:results}, we found that models trained with Eq. \ref{eq:corr-l2} as training loss outperforms the ones trained with Eq. \ref{eq:pca-l2}, therefore, we only analyze the models trained with  \ref{eq:corr-l2} in this paper.

\subsubsection{TL-DeepSSM}

\begin{figure}
    \centering
    \includegraphics[width=\linewidth]{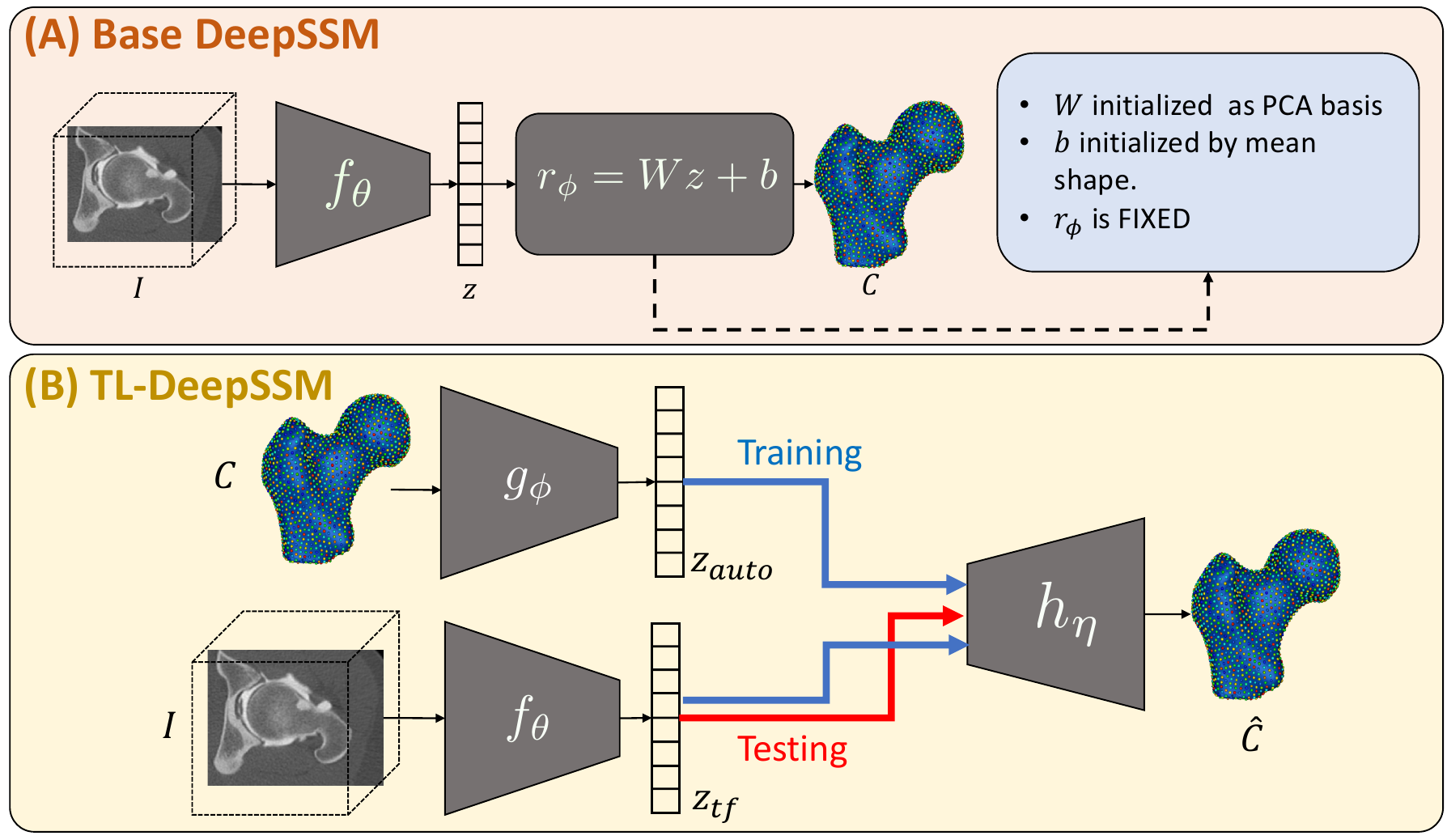}
    \caption{\textbf{DeepSSM network architecture:} (A) shows the Base-DeepSSM architecture, and (B) showcases the TL-DeepSSM architecture, the blue arrows are for training signal and the red is for the inference where we only utilize the input image.}
    \label{fig:netarch}
\end{figure}

Regression of the shape to its PCA scores is sufficient to capture all the linear shape variations in the population but will fail to capture the subtle non-linear changes that might be of interest. Although the data augmentation relies on PCA spaces, the correspondences may showcase non-linear variations (since we utilize all the PCA modes for augmentation), and the PCA score-based shape descriptor may be limiting in describing the anatomical variations. For this, we propose a variant architecture of DeepSSM that does not rely on PCA score as a shape descriptor. We take inspiration from TL network \cite{girdhar2016TLnet} architecture, which is applied for 2D to 3D reconstruction and relies on discovering a common manifold that adheres to two different representations of the same object. We utilize this idea to find a low dimensional latent representation that is \emph{common} to the PDM representation and its corresponding raw image.  The input is an image and correspondence pair given by $(I_i, C_i)$. The network architecture of the TL-variant consists of two parts: (i) the autoencoder with encoder $g_{\phi}$, and decoder $h_{\eta}$ that learns the latent code as $S_i = g_{\phi}(C_i)$ for each correspondence $C_i$, and (ii) the network that learns the latent code from the image (we call this the T-flank and it is given by $f_{\theta}$). 
This is shown in the bottom block of Figure \ref{fig:netarch}, and a detailed description of the network architecture is given in \ref{app:netdet}
The loss function of the entire architecture is given as follows:

\begin{align}
    \mathcal{L} &= \lambda_1\mathcal{L}_{auto} + \lambda_2 \mathcal{L}_{tf} \\
    \mathcal{L}_{auto} &= \frac{1}{M}\sum\limits_{i=1}^M||C_i - h_\eta(g_\phi(C_i))||^2 \\
    \mathcal{L}_{tf} &= \frac{1}{M}\sum\limits_{i=1}^M ||g_{\phi}(C_i) - f_\theta(I_i)||^2
\end{align}

We break the training routine into three parts (similar to \cite{girdhar2016TLnet}). First, we train the autoencoder using the correspondences, and hence we have $\mathcal{L} = \mathcal{L}_{auto}$ by using $\lambda_1 = 1, \lambda_2 = 0$.  Next, we train the T-flank while the  autoencoder weights are kept frozen.  The loss function is given as $\mathcal{L} = \mathcal{L}_{tf}$, by setting $\lambda_1 = 0, \lambda_2 = 1$. Finally, we train the entire model jointly with the loss function $\mathcal{L} = \mathcal{L}_{auto} + \mathcal{L}_{tf}$, and $\lambda_1 =1, \lambda_2 = 1$. For inference using a testing sample, one can directly obtain the correspondences from an image $I$ via the T-flank and decoder as $C = f_{\phi}(h_{\eta}(I))$.

\subsection{Focal Loss}
\label{sec:focal}

We showcased two different architectures of the DeepSSM model; however, both these are trained by utilizing the $\mathbb{L}$-2 norm as training loss. Some anatomy datasets need a different loss. Specifically, when the given anatomy exhibits very subtle shape variations in localized regions, i.e., the variance of the particles across the data is only significant for a small subset of correspondences. Amidst this variance imbalance across the particles, if we utilize the $\mathbb{L}$-2 correspondence loss as given in Eq. \ref{eq:corr-l2}, the network will degenerate to producing the mean shape as the output, independent of the input image. To handle such datasets, we utilize the focal loss \cite{lin2017focal}, which was initially proposed for handling class imbalance in object detection tasks but have since been extended for regression applications \cite{lu2018deep}. For a pair of input ($x$) and prediction ($\hat{x}$) the focal loss is given as:
\begin{align}
    \mathcal{L}_{focal}(x, \hat{x}) = \frac{||x - \hat{x}||^2}{1 + \exp(a \cdot (c - ||x - \hat{x}||))}
\end{align}

Here, $a$ and $c$ are hyperparameters where $a$ is a scaling factor and $c$ decides the threshold. If the particle difference is below $c$, the contribution of the particle to the overall loss is reduced. The scaling parameter $a$ is used to dampen or accentuate the loss based on the magnitude of $c$. 
 This focal loss allows for better generalization on datasets with subtle variations and prevents the model from degenerating to learning the mean shape. To incorporate focal loss with Base-DeepSSM we modify the loss in Eq. \ref{eq:corr-l2} to the following:
 \begin{align}
     \mathcal{L} = \frac{1}{M}\sum\limits_{i=1}^M\mathcal{L}_{focal}(C_i, r_{\phi}(f_\theta(I_i)) 
     \label{eq:focal-l2}
 \end{align}
For the TL-DeepSSM the focal loss is applied both for autoencoder training as well as training the latent space, and hence the loss function is modified as follows:
\begin{align}
    \mathcal{L} &= \lambda_1\mathcal{L}_{auto} + \lambda_2 \mathcal{L}_{tf} \\
    \mathcal{L}_{auto} &= \frac{1}{M}\sum\limits_{i=1}^M\mathcal{L}_{focal}(C_i, h_\eta(g_\phi(C_i))) \label{eq:autofocal}\\
    \mathcal{L}_{tf} &= \frac{1}{M}\sum\limits_{i=1}^M \mathcal{L}_{focal}(g_{\phi}(C_i), f_\theta(I_i)) \label{eq:latfocal}
\end{align}

The hyperparamenter $c$ is estimated using a heuristic. We compute the absolute difference per point for each sample in the training data with respect to the mean shape, i.e., $S(i,j) = ||C^j_i - \mu^j||$, for $i^{th}$ sample and $j^{th}$ correspondence. We choose $c$ as the score corresponding to the 90 percentile value. We set the scaling hyperparameter $a$ to 10 in all our experiments for particle based focal loss (eq. \ref{eq:focal-l2} and eq. \ref{eq:autofocal}), as in most cases $c$ lies between 1-2. 
The TL-DeepSSM with the focal loss, has focal loss applied at two different places and each of them will have its own a and c parameters. First, for the correspondences autoencoder, whose a and c parameters are discovered similarly to that of the base-DeepSSM, as the loss is applied on the correspondences. Second, for the latent space, we use the trained autoencoder and the latent space for the training data to get an estimate of c as described above, here the estimates of c are in orders of 10 for all dataset and hence we set our a as 1 for all cases.



\section{Results}
\label{sec:results}
In this section, we describe the application of DeepSSM on three different datasets provided in separate subsections. We also describe the specific training methodology and hyperparameters used for each dataset in their respective sections. Some training and implementation details are common across experiments; these are compiled together in  \ref{app:netdet} for clarity. 


\textbf{Evaluation: } In the upcoming subsections, we evaluate DeepSSM to showcase that it performs similarly to a state-of-the-art PDM model by comparing the dense correspondences. We re-iteratate here that for a state-of-the-art PDM we need to segment the raw images and process them before feeding them into the PDM algorithm. For each dataset, we evaluate both Base-DeepSSM and TL-DeepSSM with and without focal loss. Error is captured via average relative mean squared error (RMSE) between the predicted 3D correspondences and state-of-the-art. We compute this by averaging the RMSE for $x$,$y$, and $z$ coordinates as follows:
\begin{align}
    RMSE = \frac{1}{3}(RMSE_x + RMSE_y + RMSE_z)
\end{align}
Where, for N 3D correspondences, $RMSE_x = \sqrt{\frac{||C_x - \hat{C}_x||_2^2}{N}}$, and similarly for y and z coordinates. We also compute the RMSE error for each correspondence point, i.e., $RMSE_i = \sqrt{\frac{||C^i_x - \hat{C}^i_x||_2^2+ ||C^i_y - \hat{C}^i_y||_2^2+ ||C^i_z - \hat{C}^i_z||_2^2}{3}}$. The mean and standard deviation of these per-point RMSE are visualized as a heatmap on top of the mean shape. Per-point RMSE informs us how well DeepSSM is able to model different local anatomy. Along with the RMSE, we also report the surface-to-surface distance between the ground truth mesh and the mesh reconstructed from the particles predicted by DeepSSM. The surface-to-surface distance gives a more accurate portrayal of  particles adhering to the shape and gives insight into whether or not they can be used for facilitating anatomy segmentation. 
Additionally, to validate the efficacy of DeepSSM as an estimator of shape representation, we use the DeepSSM correspondences for different downstream analysis applications and compare the results to that of a state-of-the-art PDM. These downstream applications are different for each dataset and are described separately in corresponding subsections.

\textbf{Inference Time: } One of the contribution of the DeepSSM model is fast inference on new samples, i.e., using a trained model to predict correspondences from a new image. Comparison of the inference time using DeepSSM and state-of-the-art PDM, \emph{ShapeWorks} \cite{cates2017shapeworks} for a single scan is shown in Table \ref{tab:time}, and we see the drastic improvement that DeepSSM provides. The DeepSSM time provided are for Base-DeepSSM, and should be noted that the TL-DeepSSM has very similar inference times. In these results ShapeWorks was executed on a system with 192 Intel Xeon Gold 6252 CPU @ 2.10GHz(HT) cluster, with multi-threading. The DeepSSM results are computed on the same machine utilizing the CPU for a fair comparison. These results are only reporting the inference time and ignores processing time required for the input, which is very high for Shapeworks (or any PDM) as it includes segmentation and heavy pre-processing, and is zero for DeepSSM, since it acts directly on images.  



\begin{figure}[!h]
    \centering
    \includegraphics[width=0.9\textwidth]{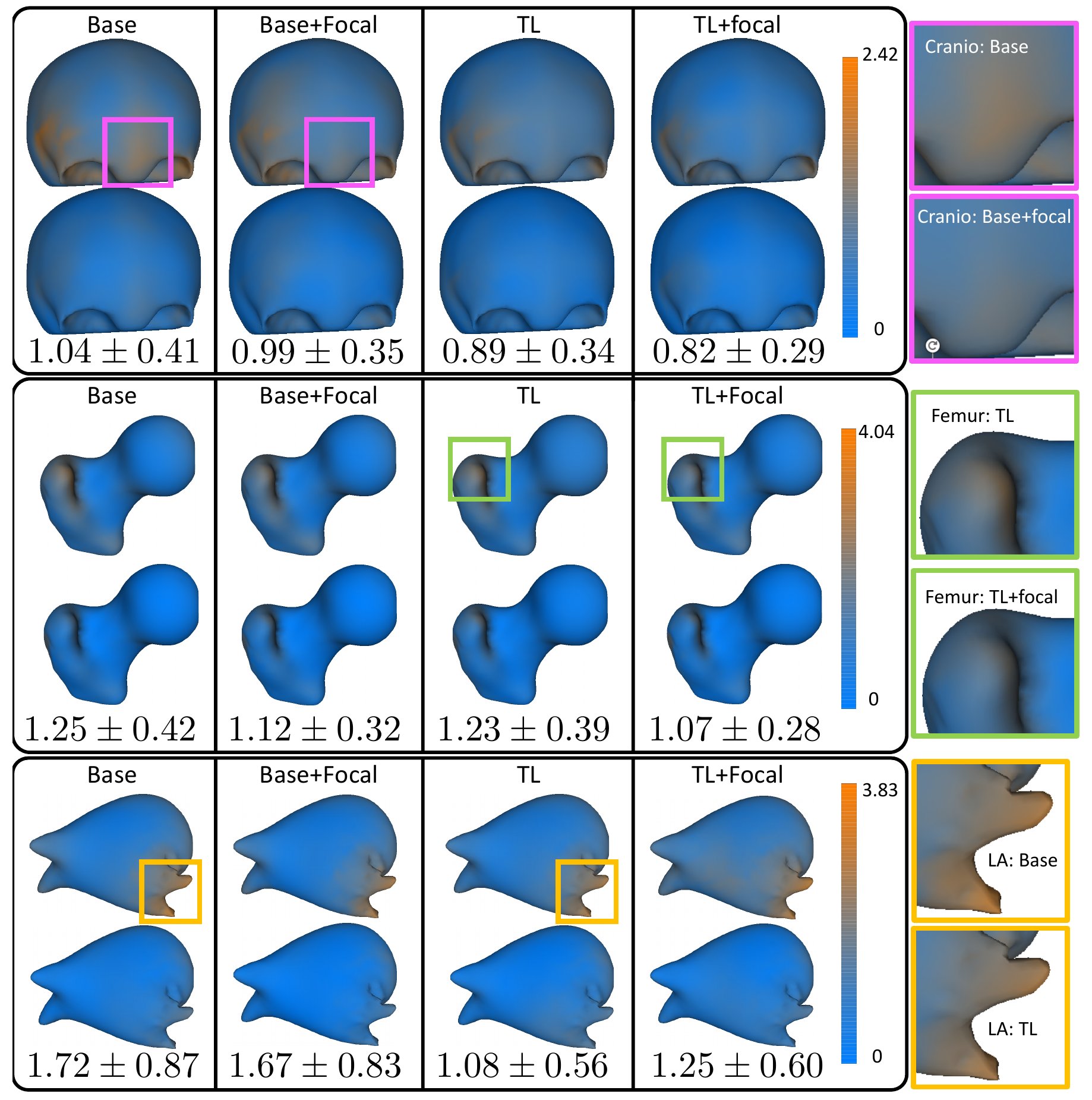}
    \caption{\textbf{Point-wise RMSE errors.} Each block represent the mean (top-figure) and standard deviation (bottom-figure) for a given dataset and DeepSSM Variant. Additionally, the average RMSE error over the test data is given by the number at the bottom. All the units are in millimeters, and 'focal' denotes the model trained with the focal loss. On the right of the table are zoomed in versions of local anatomy captured in respective colored boxes.}
    \label{fig:rmse_plots}
\end{figure}

\begin{table}[!h]
\centering
\begin{tabular}{|l|l|l|l|}
\hline
\textbf{Data}  & \textbf{$N$}    & \textbf{ShapeWorks} \cite{cates2017shapeworks} & \textbf{DeepSSM}             \\ \hline
\textbf{Metopic}   & 2048 & 39 minutes           & \textbf{0.56} seconds \\ \hline
\textbf{LA}        & 1024 & 31 minutes          & \textbf{0.28} seconds \\ \hline
\textbf{Femur}     & 1024 & 65 minutes           & \textbf{1.25} seconds \\ \hline
\end{tabular}

\caption{\textbf{Inference time comparison: } Inference time for a single image for ShapeWorks \cite{cates2017shapeworks} versus Base-DeepSSM, where N denotes the number of particles. }
\label{tab:time}
\end{table}

\subsection{Metopic Craniosynostosis}

Craniosynostosis is a morphological defect that affects infants, where one of the sutures of the skull is fused prematurely, and the subsequent brain growth leads to abnormal head shapes. In severe cases, craniosynostosis can lead to increased intracranial pressure, causing symptoms such as headaches, drowsiness, vision changes, and more. It can also cause several neurological problems. Metopic craniosynostosis is craniosynostosis where the metopic suture (also called frontal suture) is fused prematurely. It leads to an unnaturally triangular head shape, a morphological symptom known as trigonocephaly 

\cite{kellogg2012interfrontal}.  Compensatory growth in metopic craniosynostosis also causes expansion at the back of the skull. Metopic craniosynostosis is difficult to identify with CT scans as the suture fuses typically at an early age, making it difficult to ascertain whether or not the suture fused prematurely. The current diagnosis protocol includes subjective observation of the head shape. The diagnosis affects the subsequent treatment protocol, namely to perform corrective craniofacial surgery or not. This treatment is risky by nature and requires an objective measure of the severity of the condition. This objective severity measure for craniosynostosis will guide the surgeons to make a better-informed diagnosis. In this section, we will utilize DeepSSM to construct a shape model on cranial data. The end goal is to quantify the deviation of a given shape diagnosed with metopic craniosynostosis from the population of normal cranial shapes. 

\textbf{Data Description and Processing:} Cranial CT scans of infants between 5-15 months of age were acquired at UPMC Children’s Hospital of Pittsburgh. A subset of 28 scans were diagnosed with metopic craniosynostosis by a board-certified craniofacial plastic surgeon utilizing the CT images as well as a physical examination. The remaining 92 scans are CT scans of trauma subjects acquired as a by-product of standard clinical practice; however, these scans do not exhibit any deformities and can be treated as a set of control patients. All the data is IRB approved and acquired by utilizing standard de-identification and HIPAA protocols. Along with the CT scans, we have corresponding segmentations of the cranium. We separate this dataset into 105 training and 15 testing images. We process images and segmentations for the training data and utilize the \emph{ShapeWorks} \cite{cates2017shapeworks} package to generate the initial PDM on the data. To satisfy the GPU memory requirements, all of the images are downsampled isotropically by a factor of 4. This results in volume size of $69 \times 51 \times 66$ and resulting voxel spacing of 4mm in each direction. The PDM consists of 1024 3D points and uses 20 PCA components (capturing 99\% of variability). We perform data augmentation and generate 5000 additional data points. The total 5000 augmented plus 105 original training images are divided into training and validation set with a 90\%-10\% split. 

\textbf{Training Specifics: } We train the Base-DeepSSM with 20 PCA components (explaining 99\% of data variability), using loss on correspondences (Eq. \ref{eq:corr-l2}) for 100 epochs with early stopping with respect to validation loss. Similarly, we also train the Base-DeepSSM with focal loss as given by Eq. \ref{eq:focal-l2} for 100 epochs. Finally, we train a TL-DeepSSM with a bottleneck of 64 dimensions with and without the focal loss. We train the autoencoder for 5000 epochs, the T-flank for 100 epochs, and the joint model for 50 epochs. The values of the hyperparameter $c$ for the focal loss was discovered by the heuristic described in Section \ref{sec:focal}, and is 1.09 for particles (in base and TL DeepSSM) and 10.9 for TL latent space. 

\begin{figure}
    \centering
    \includegraphics[width=\textwidth]{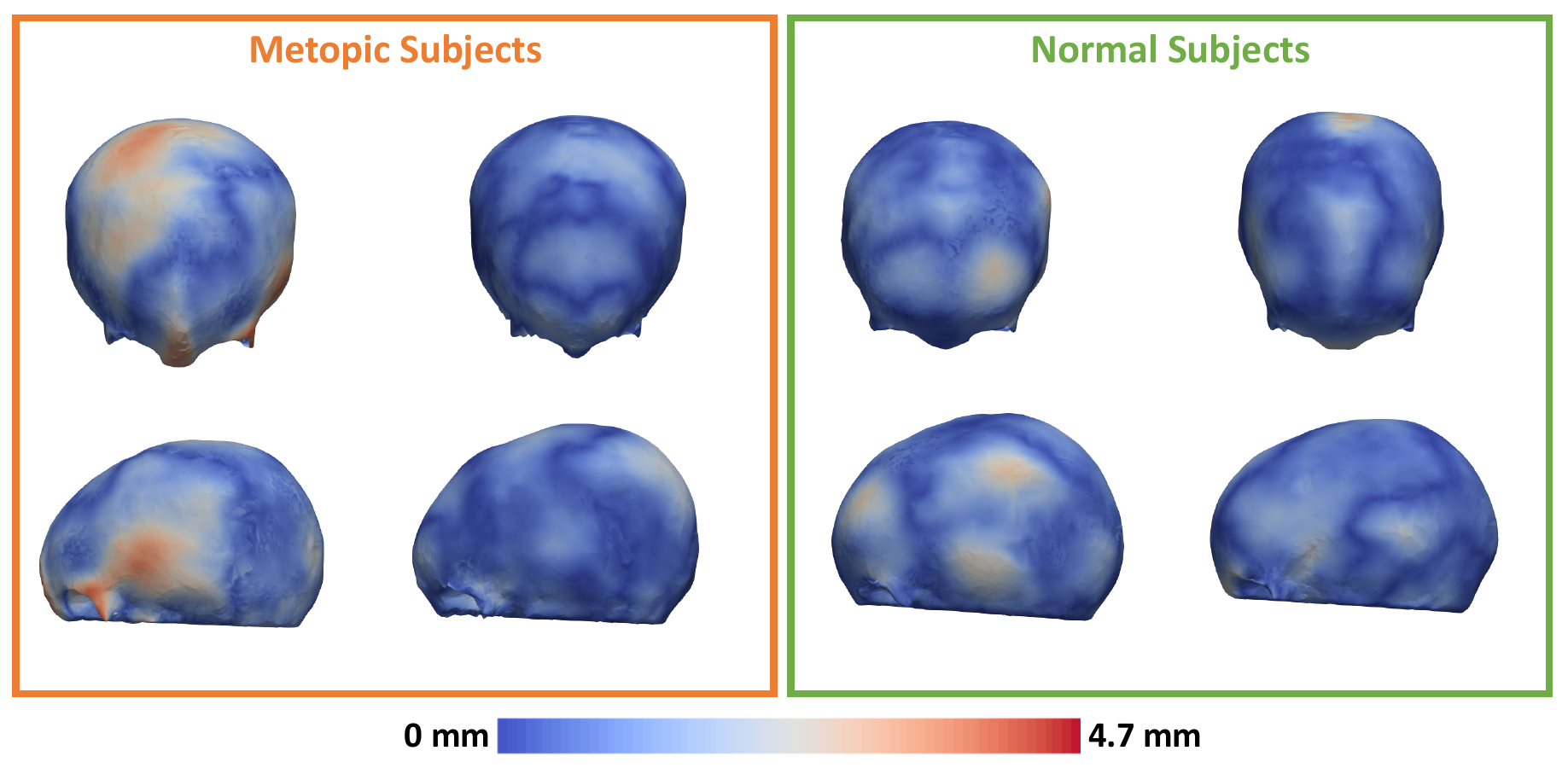}
    \caption{\textbf{Qualitative results on metopic craniosynostosis data.} Surface-to-surface distance between the meshes reconstructed from Base-DeepSSM model trained without focal loss and the ground truth meshes. The evaluation is performed on normal and metopic scans from testing set.}
    \label{fig:cranio_qual}
\end{figure}

\begin{figure}
    \centering
    \includegraphics[width=\textwidth]{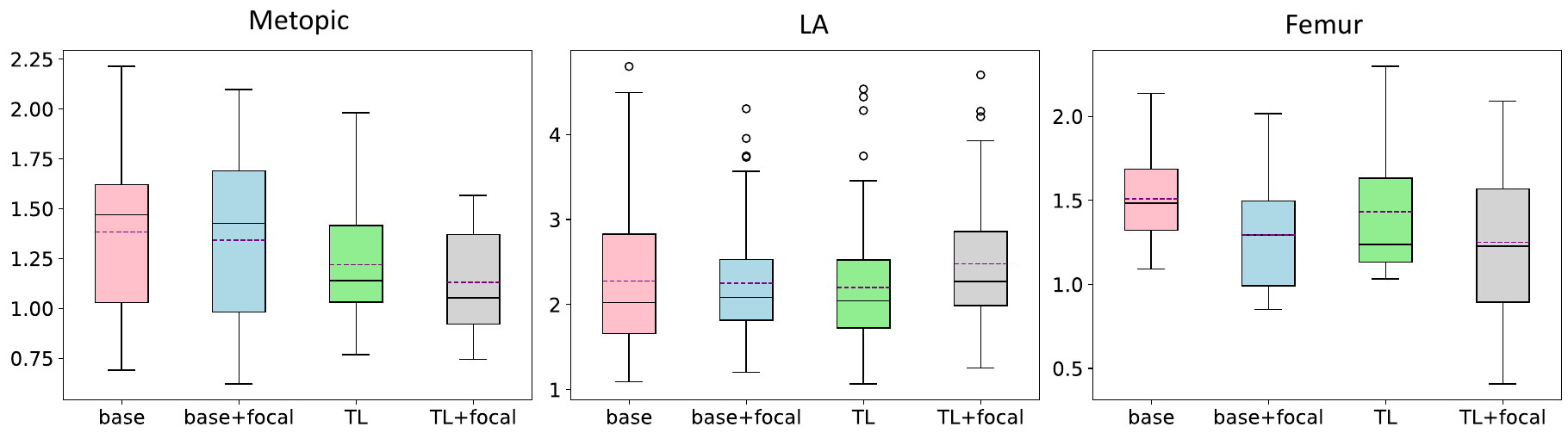}
    \caption{\textbf{Surface-to-surface distance box-plots}. Box plots showcase the surface to surface distance values between ground truth mesh and mesh obtained from DeepSSM predictions across all test samples. The y-axis is the surface-to-surface distance in millimeters, and 'focal' denotes models trained with focal loss. The black lines in each box-plot mark the median value and the purple line marks the mean value. }
    \label{fig:boxplts}
\end{figure}

\textbf{Evaluation and Analysis: } The average RMSE and the per-point RMSE are given in Figure \ref{fig:rmse_plots}, and the average surface-to-surface distance for each test sample is provided via box-plots in Figure \ref{fig:boxplts}. These metrics, namely RMSE, per-point RMSE, and surface-to-surface distance are computed as described in the beginning of the Section \ref{sec:results}. Qualitative results are showcased in Figure \ref{fig:cranio_qual}, the heatmap is the surface-to-surface distance between the original mesh and the mesh reconstructed from the Base-DeepSSM (without focal loss) predicted particles. From Figure \ref{fig:rmse_plots} we can see that the prediction error is centered around the eye sockets or frontal ridge, however the average RMSE is well within the sub-voxel margin, as the input images have spacing of 4mm. We can see that both normal and metopic test scans capture the shape well and with a sub-voxel accuracy in most areas. The left-most metopic scan in Figure \ref{fig:cranio_qual} is the result with the largest error, and the maximum error is localized in the region near the orbital rim. The oribital rim is a sharp and thin structure, and the original PDM on which DeepSSM is trained does not provide very accurate correspondences to the region. These inaccuracies persists in DeepSSM training and the trained model also exhibits this error in the localized area. This is a data problem, and the neural network does not learn the structures that are not well expressed by data. The improvement provided by focal loss is minimal both quantitatively as well as qualitatively, however, we can see from Figure \ref{fig:rmse_plots} that the distribution of RMSE error is more spread out and less localized in specific regions. Because focal loss does not provide significant improvement on this dataset, we the Base-DeepSSM and TL-DeepSSM trained without focal loss for the downstream application and analysis.

\textbf{Downstream Task -- Severity Quantification: }
Our end goal is to utilize DeepSSM to provide clinicians with an end-to-end tool where they can input an image and acquire a severity score. 

The severity score is designed to measure the abnormality of a subject with respect to the normal population (samples with a negative diagnosis for metopic craniosynostosis). We utilize an unsupervised learning method that models the variations in the normal population and learns the deviation between a test image and this population \cite{tao2020unsupervised}. The method employs Mahalanobis distance/Z-score, which uses the variance in the normal population to normalize the multidimensional difference of the target subject. It can be regarded as a Z-score for multiple dimensions and effectively scales up rare characteristics in the shape statistics of the normal population. 

The sample number is usually limited and much lower than the shape dimension; we account for the variability outside the sample subspace using Probabilistic Principal Component Analysis (PPCA). PPCA estimates the full rank covariance $S$ as given by Equations \eqref{eq:ppca_mle} and \eqref{eq:ppca_cov}. Here, $D$ is the shape dimension, $L$ is the subspace dimension, $\Bar{C}$ is the empirical mean, $U$ are the columns of eigenvectors and $ \Lambda $ is the diagonal matrix of eigenvalues of $  \frac{1}{N} \sum_i (C_i - \Bar{C}) (C_i - \Bar{C})^T $. We can use the eigendecomposition $ U_S \Lambda_S U_S^T = S $ to whiten a shape deviation, project back to the shape space and get the point-wise Mahalanobis Distance $ \Tilde{C_i} $ which explains the contributions across the shape domain to the final severity score (as shown in Equation \eqref{eq:ppca_whiten}).

\begin{align}
    \frac{1}{N} \sum_i (C_i - \Bar{C}) (C_i - \Bar{C})^T & = U \Lambda U^T, {\sigma}^2 = \frac{ \sum_{i=L+1}^{D} \Lambda_i }{D-L}
    \label{eq:ppca_mle}\\
    S &= U_{[:L]} \Lambda_{[:L]} U_{[:L]}^T + {\sigma}^2 I
    \label{eq:ppca_cov} \\
    \Tilde{C_i} &= (C_i - \Bar{C})^T U_S \Lambda^{-\frac{1}{2}} U_S^T
    \label{eq:ppca_whiten}
\end{align}

Evaluation result for the severity scores is shown in Table \ref{tab:cranio_table}. For each shape model, we fitted PPCA (with 95\% variance subspace) on 72 control scans and calculated the severity scores for the remaining 48 scans (these include 28 metopic and 20 normal scans). The latter group is provided with aggregated expert ratings of the degree of metopic craniosynostosis, collected through an online portal with 36 craniofacial experts. 
We displayed the 3d segmentation and asked the physicians to estimate the deformity on a 5 point Likert scale. We modeled the expert ratings by Latent trait theory \cite{uebersax1993latent, muthen1998statistical} which parameterized potential raters' bias and inconsistency and obtained the aggregated latent severity using Maximum Likelihood Estimation.

We use both Pearson and Spearman correlation coefficients to measure the similarity between the predicted severity scores and the discovered latent trait using the expert ratings. The results presented in Table \ref{tab:cranio_table} suggest that both base- and TL-DeepSSM have severity scores that are highly correlated with the expert consensus. In addition, we also calculate the Area Under Curve (AUC) using the binary diagnosis to evaluate the efficacy of the predicted severity scores for classification. The AUC also suggest that DeepSSM based severity is a good indicator to identify scans with metopic craniosynostosis. Both AUC and correlation scores for the DeepSSM variants are comparable to results using \emph{ShapeWorks} \cite{cates2017shapeworks}, hence verifying that DeepSSM can perform comparably with state-of-the-art PDM at this downstream task.

\begin{table}[]
\centering
\begin{tabular}{l|l|l|l|}
\cline{1-4}
    \multicolumn{1}{|l|}{ \textbf{Metric \&  Model}}  & Base & TL & ShapeWorks \cite{cates2017shapeworks} \\ \hline
    \multicolumn{1}{|l|}{\textbf{PearsonR}}  & 0.82  & 0.82 & 0.85 \\ \hline
    \multicolumn{1}{|l|}{\textbf{SpearmanR}}  & 0.80  & 0.81 & 0.83 \\ \hline
    \multicolumn{1}{|l|}{\textbf{AUC}}  & 0.946 & 0.98 & 0.932 \\ \hline
\end{tabular}
\caption{\textbf{Evaluation of estimated severity scores:}  Correlation and AUC of the severity measure using different shape models with respect to the expert ratings and higher numbers are better. The shape models used are state-of-the-art PDM ShapeWorks \cite{cates2007shape}, the base and TL-DeepSSM trained without the focal loss.}
\label{tab:cranio_table}
\end{table}

\begin{figure}[!h]
    \centering
    \includegraphics[width=0.85\textwidth]{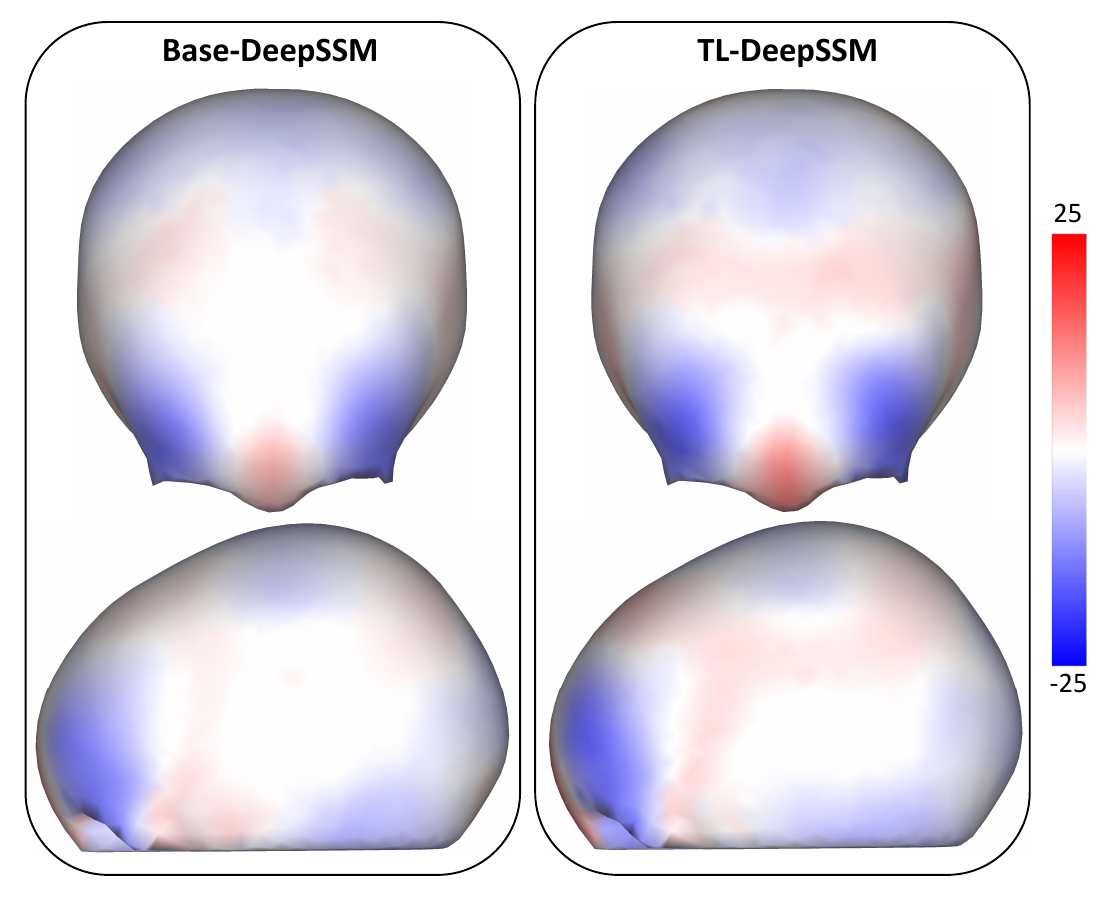}
    \caption{\textbf{Average Point-wise Mahalanobis Distance between metopic and normal control data.} The upper figure shows  the top view and the bottom figure shows the side view. The heatmap shows the magnitude of the deviation in both directions, with blue being inward motions and red being outward.}
    \label{fig:cranio_point_maha}
\end{figure}

We displayed the point-wise Mahalanobis distance across the shape domain as is shown in Figure \ref{fig:cranio_point_maha}. The point-wise Mahalanobis distance is obtained by averaging the whitened deviations of all metopic scans.  It is then projected onto the surface normals and visualized using the heap map on the top of the mean control shape in both top and side views. The heat maps show that the center of the frontal bone protrudes while the sides shrink; this morphological deviation showcases trigonocephaly, which is a clinically expected observation. Therefore, DeepSSM shape model captures the characteristic deformation pattern for metopic craniosynostosis. 

\subsection{Cam-type Femoroacetabular Impingement}
\label{sec:cam}

Cam-type femoroacetabular impingement (cam-FAI) is characterized by an abnormal bone growth of the femoral head. The abnormal bone growth causes excessive friction between the hip socket(part of the large pelvis bone) and the femoral head \cite{harris2013cam}. This friction can cause eventual joint damage leading to pain and restricted movement, which is also a leading cause of hip osteoarthritis. Modeling the shape from a joint population of femurs diagnosed with and without CAM-FAI  can characterize the local femoral head deformity (cam-lesion) with respect to the femoral shape statistics of healthy subjects. Such a characterization can provide surgeons an anatomical map of the deformity and therefore driving the surgical intervention. It can also help understand the pathology more intuitively, and correct for local abnormal morphology and preserve normal bone function. We intend to utilize DeepSSM models to achieve this characterization and compare it to the state-of-the-art PDM models, showcasing the efficacy of utilizing DeepSSM for such tasks while mitigating the need for segmentation and added pre-processing.

\textbf{Data Description and Processing: } For training the DeepSSM models, we have 49 CT images of the femur bone. Out of these 49 CT, 42 subjects are healthy and not diagnosed with any morphological abnormality in the femur bone, and we call these control scans. The remaining 7 scans are diagnosed with CAM-FAI. For the initial shape model, we use \emph{ShapeWorks} \cite{cates2017shapeworks} and generate 1024 correspondences on these shapes. Next, we apply data augmentation on these 49 scans producing 5000 additional data points. This combined 5049 scans are split into 90-10\% split for training and validation. We utilize 20 PCA modes (capturing ~99\% variability) for training the DeepSSM. In addition, to the training and validation data, we also have 7 controls and 2 CAM-FAI scans reserved for testing the DeepSSM model. Each image is downsampled by a factor of 4 to fit the GPU memory requirements easily. This downsampling results in the 3D volumes with dimensions of $65 \times 46 \times 58$ voxels, with isotropic voxel spacing of 2mm.


\textbf{Training Specifics: } We train the femur model for the Base-DeepSSM and TL-DeepSSM with and without the focal loss. We train the baseline DeepSSM with 20 PCA modes without focal loss via Eq. \ref{eq:corr-l2} and with focal loss by Eq. \ref{eq:focal-l2}. We train each of these models for 50 epochs with early stopping based on the validation loss.  Finally, we train a TL-DeepSSM with and without the focal loss, where we train the autoencoder for 10000 epochs followed by training the T-flank for 50 epochs and joint training for 25 epochs. The values of the hyperparameter c for the focal loss was discovered by the heuristic described in Section \ref{sec:focal}, and is 1.32 for particles (in base and TL DeepSSM) and 6.3 for TL latent space.

\begin{figure}[!h]
    \centering
    \includegraphics[width=\textwidth]{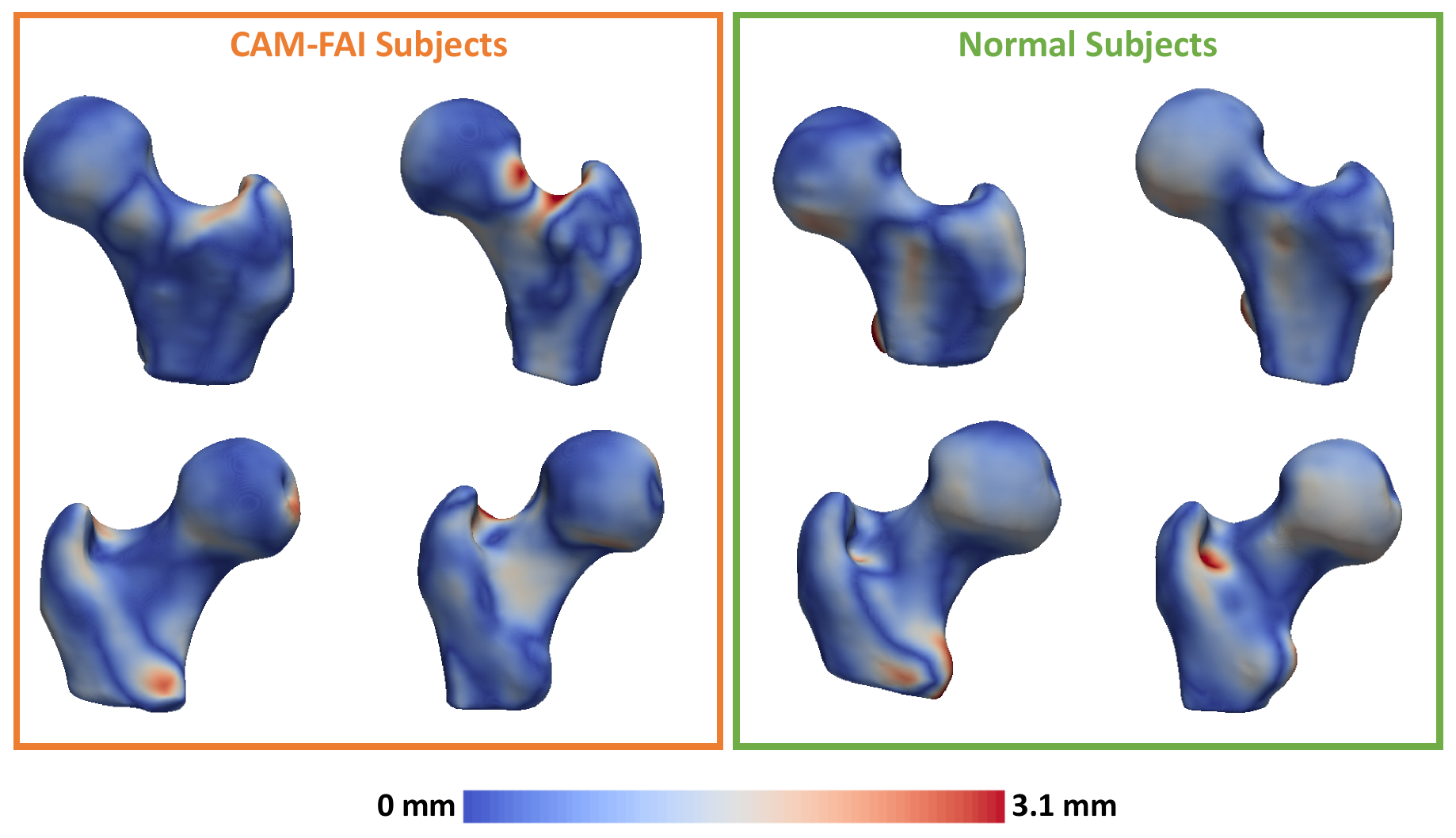}
    \caption{\textbf{Qualitative Results on Femurs: } These shows four randomly chosen samples from the test dataset 2 from control/normal set and 2 CAM examples. Each column is a single example shown in two different views. Each mesh is the original shape with the heatmap representing the surface-to-surface distance between the original mesh and mesh constructed from DeepSSM predicted correspondences. These are computed using the Base-DeepSSM model with focal loss. }
    \label{fig:femur_qual}
\end{figure}

\textbf{Results and Analysis: }  The average and per-point RMSE are given in \ref{fig:rmse_plots}. Surface-to-surface distance is computed between the mesh reconstructed from the DeepSSM predicted correspondences and the mesh coming from ground-truth segmentation of the femur bone, and is given in Figure \ref{fig:boxplts}.  We have two key observations, first, we notice that focal loss improves the RMSE scores marginally, however this improvement is crucial in capturing the subtle variability in femur anatomy. We also see from Figure \ref{fig:rmse_plots} that the models with focal loss have a more diffused loss and is less persistent in greater trochanter area. Second, we notice that the TL variant outperform the Base-DeepSSM but not by a huge margin indicating that the PCA scores as a shape descriptor captures most of the variability. In this dataset we notice that the models trained without the focal loss degenerate to producing a similar predictions in areas which have large variations in shape and as they are very limited in number the loss is still low in value. However, this is meaningless as the areas with larger variance are the most important in characterizing the anatomy and we want DeepSSM to capture them.
Qualitative results using the Base-DeepSSM with focal loss are shown in Figure \ref{fig:femur_qual}, it showcases the surface-to-surface distance on testing images (both controls and CAM-FAI). 

We notice that the error on the lesser trochanter and the CAM lesion is relatively high in some cases; this is due to two reasons. First, the plane at which the femurs are chopped off is arbitrary, and in some cases (both in training and testing images), the entire lesser trochanter is not captured. This inconsistency in cutting planes introduces a high and uncertain shape variation in that region that is hard for the network to learn. This is a data issue, and can be fixed by standardizing the scans fed as an input to the model. Second, the error in CAM lesion occurs due to data imbalance, i.e., there are many more controls than the CAM-FAI diagnosed femurs. This results in extreme cases of CAM (such as the second CAM case in Figure \ref{fig:femur_qual}) not completely being captured by the DeepSSM model. DeepSSM model can only encode the variations that are present in training data, and if the dataset were larger and well balanced between two classes, the performance of the model would increase. 

\begin{figure}[!h]
    \centering
    \includegraphics[width=0.75\textwidth]{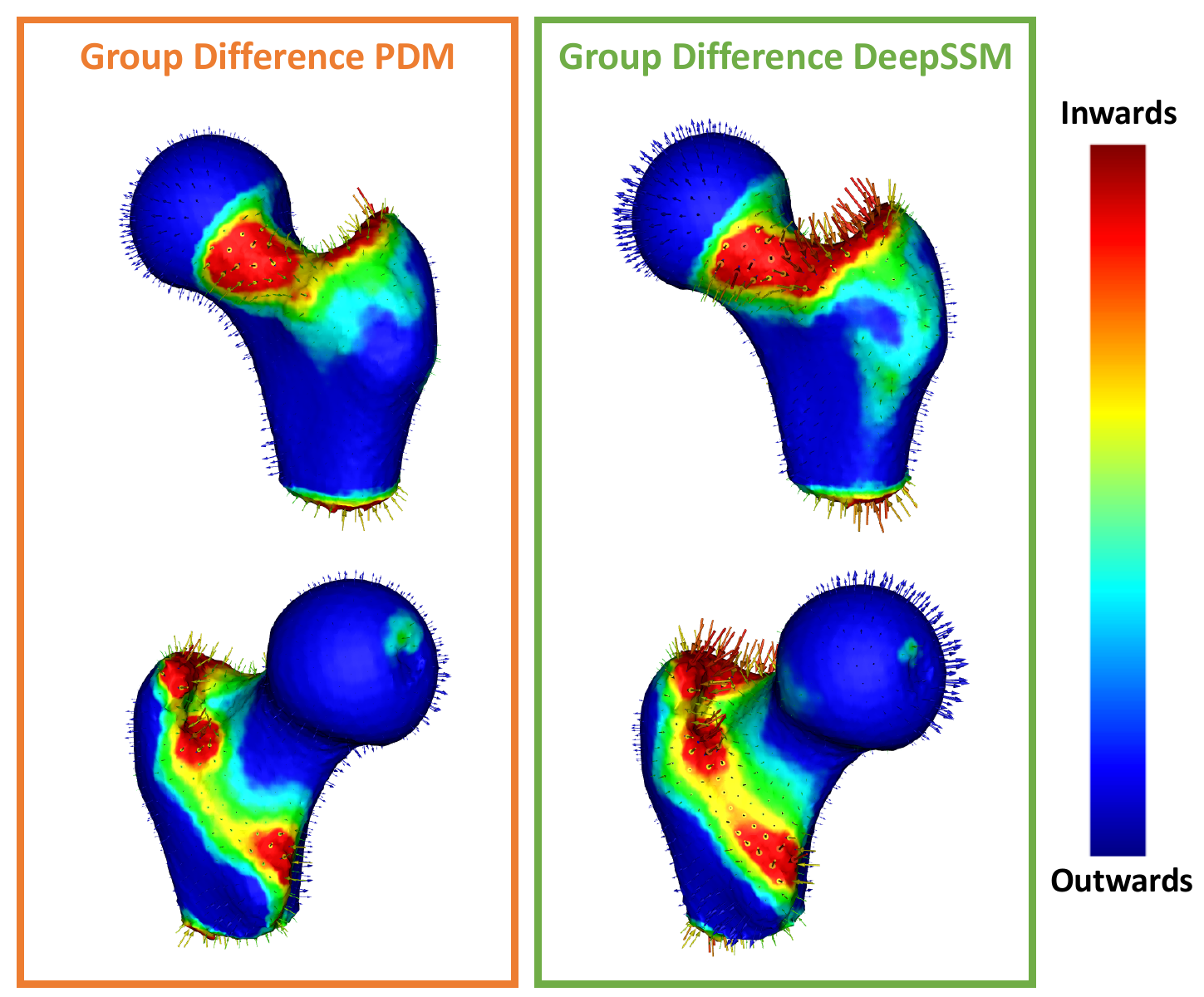}
    \caption{\textbf{Femur shape group difference between CAM-FAI and controls.} The difference $\mu_{cam} - \mu_{normal}$ and is projected on $\mu_{cam}$, the arrows denote the direction of the particle movement and the heatmap showcases the normalized magnitude.}
    \label{fig:femur_gd}
\end{figure}

\textbf{Downstream Analysis -- Group Differences:} As mentioned earlier, it is of interest to the clinical community to capture the statistical morphological difference between the CAM-FAI shape and the representative control femur bone shape. This characterization is better represented by models trained with focal loss and hence, for this downstream analysis we use base- and TL-DeepSSM trained with the focal loss.
We construct two groups, one for controls and the other for pathology (cam-FAI), and compute the difference between their means ($\mu_{normal}$ and $\mu_{cam}$) and showcase this difference on a mesh. This is termed as a \emph{group difference} \cite{harris2013cam}. We compute this group difference utilizing the DeepSSM predicted particles, as well as via \emph{ShapeWorks} PDM model. The entire data is used (both testing and training) for these group differences, and the figures are shown in Figure \ref{fig:femur_gd}. Each group difference showcases the difference going from pathological mean shape to control the mean shape and is overlayed on top of the mean pathological scan. We can see that the group differences for both the state-of-the-art PDM model and DeepSSM are very similar. Therefore showcasing that DeepSSM can be used to get correspondences without undergoing heavy pre-processing and segmentation steps, and these correspondences can be used to characterize the CAM deformity.

\begin{figure}
    \centering
    \includegraphics[width=\textwidth]{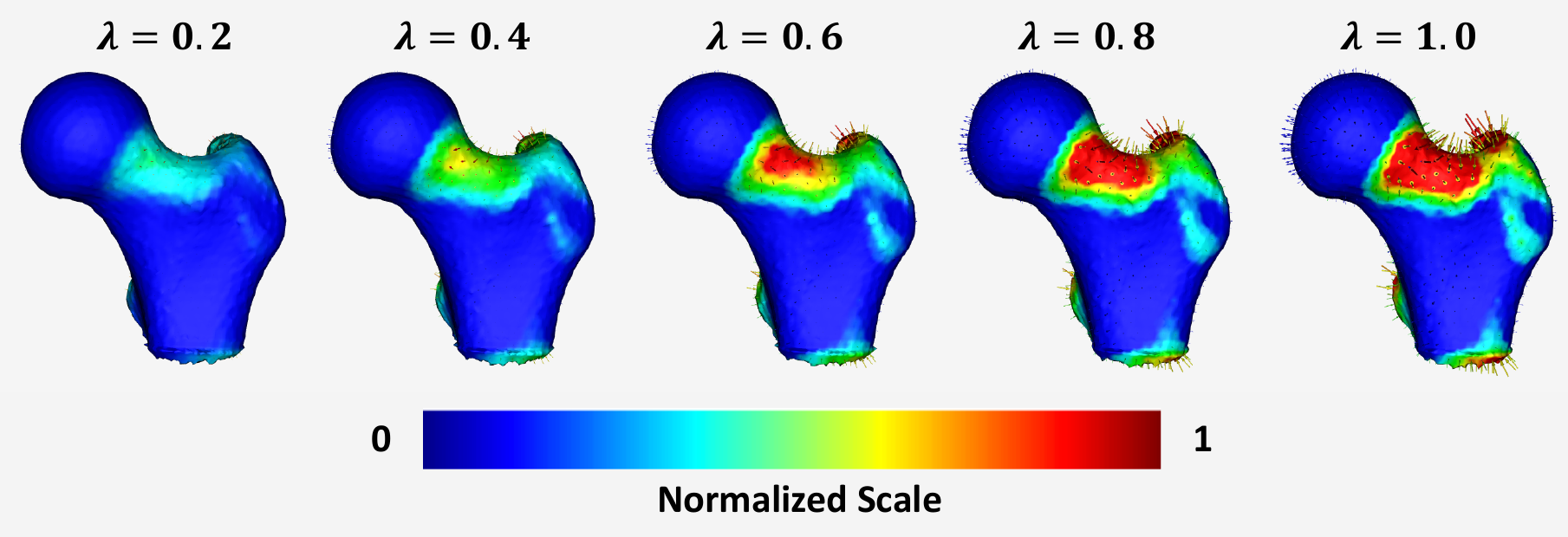}
    \caption{\textbf{TL-DeepSSM latent swim}: Samples taken in latent space along the line between mean normal and mean pathological latent vector, and its difference from the normal shape is overlayed over each sample.}
    \label{fig:lda_femur}
\end{figure}

TL-DeepSSM predicts a latent space that captures non-linear variations, but at the same time, can also capture the morphological differences between normal and CAM-FAI femoral shape in its latent space. To show this we move along the line connecting mean normal ($\mu_{norm}$) and mean pathological ($\mu_{cam}$) \emph{in the latent space}, and then at each sample we utilize the decoder($f_\phi$) to get the corresponding correspondences, i.e $C_\lambda = f_\phi((1 - \lambda)\mu_{norm} + \lambda\mu_{cam})$. Figure \ref{fig:lda_femur} shows the difference between each of these samples and mean normal shape, i.e., $C_0 - C_\lambda$, overlaid on $C_\lambda$. We observe the clinically expected outcome, as femur anatomy smoothly exhibiting inward motion around the CAM lesion as $\lambda$ increases. 

\subsection{Atrial Fibrillation} 
\label{sec:la}

Atrial Fibrillation (AF) is irregular heartbeat (arrhythmia) that can lead to several problems such as stroke, blood clots, and even cardiac arrest. Catheter ablation is a therapeutic procedure that is commonly utilized to treat AF; however, many patients can exhibit a recurrence of AF after ablation. This AF recurrence is demonstrated to be related to the shape of the left atrium \cite{bieging2018left, bhalodia2018endtoend}. We utilize DeepSSM models and their predicted shape descriptors to validate this hypothesis and estimate AF recurrence based on the shape of the left atrium.

\textbf{Data Description and Processing: } 
For the left atrium (LA) experiments, we utilize 206 late gadolinium enhancement (LGE) MRI images from patients that have been diagnosed with atrial fibrillation (AF). These scans are acquired after the first ablation.  Of the 206 scans, 122 are used in training, and additional 84 scans are kept as testing set curated carefully, ensuring even distribution of intensity and shape variation in the training versus testing data. A PDM of 1024 correspondences is generated using ShapeWorks \cite{cates2017shapeworks} for the training set, then using this as a base shape model, correspondences are also found for the test set for evaluation. We extract 19 modes from the PDM via PCA, which captures ~95\% of the shape variability. We then run data-augmentation on the training set to create 5000 additional samples, 80\% of which are used in training and 20\% in validation. For training purposes, the MRIs are downsampled from $235 \times 138 \times 175$ (0.625mm isotropic voxel spacing) with a factor of 2, resulting in images of size $118 \times 69 \times 88$ (1.25mm voxel spacing).

\textbf{Training Specifics: } We train the Base-DeepSSM with and without the focal loss for 100 epochs with 19 PCA scores that capture (~95\%) data variability. We use early stopping to evaluate performance on the best epoch. For the TL-DeepSSM (with and without the focal loss), we train the autoencoder for 10000 epochs, followed by 100 epochs to train the latent space and 25 epochs for joint training. The autoencoder bottleneck of 64 was chosen for TL-DeepSSM. The hyperparameter c used in focal loss is 2.54 for the particle loss (base and TL-DeepSSM), and is 13.76 for TL-DeepSSM latent space loss.

\begin{figure}
    \centering
    \includegraphics[width=\textwidth]{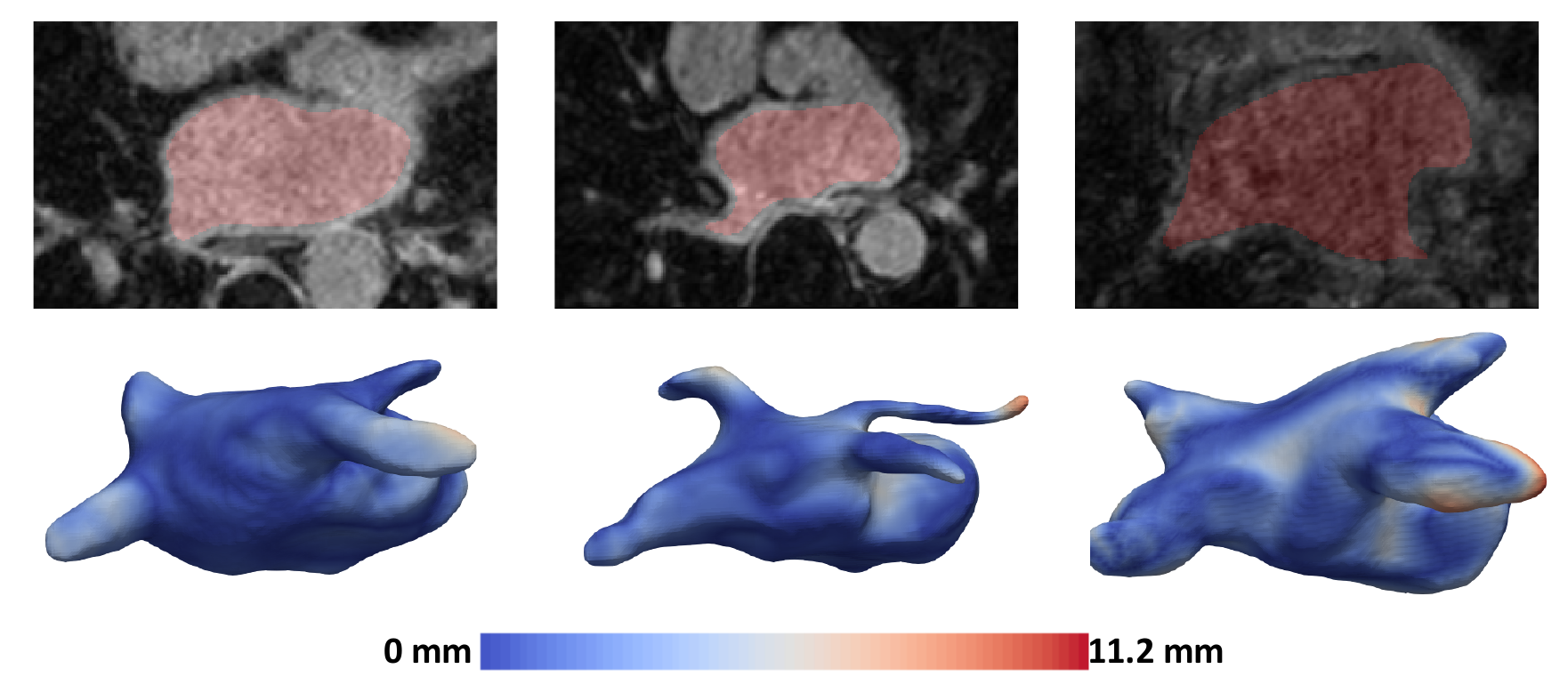}
    \caption{\textbf{Qualitative Results on Left Atrium: } These shows three samples from the test dataset, with varying degree of degradation in the corresponding LGE image. Each mesh is the original shape with the heatmap representing the surface-to-surface distance between the original mesh and mesh constructed from DeepSSM predicted correspondences. These are computed using the TL-DeepSSM model.}
    \label{fig:la_qual}
\end{figure}

\textbf{Results and Analysis: } This is a challenging dataset, firstly in terms of the intensity variations present in the LGE images (Figure \ref{fig:la_qual}), with some images being significantly degraded and noisy. Secondly, in terms of shape variation, the left atrium anatomy exhibits natural clustering based on the shape of the atrium and based on the different numbers of pulmonary veins and their orientation \cite{ho2012left}. Additionally, determining where the pulmonary veins should end in a segmentation is arbitrary across samples and across annotators. Due to this highly non-linear shape distribution, the TL-DeepSSM outperforms the other variants considerably. The RMSE results are shown ion Figure \ref{fig:rmse_plots}, and the surface-to-surface results are shown in Figure \ref{fig:boxplts}. Similar to the craniosynostosis dataset, here the results with and without the focal loss does not make much difference quantitatively as well as qualitatively.
The TL-network provides considerable improvement over the average RMSE as seen in Figure \ref{fig:rmse_plots}, with majority of the error still being concentrated around the pulmonary veins similar to that of Base-DeepSSM. Qualitative results are showcased in Figure \ref{fig:la_qual}, which depicts surface-to-surface distance between the ground truth mesh and the mesh reconstructed from the TL-DeepSSM predicted particles. We notice that the surface-to-surface distance shown in these scans is as large as 11 mm, which ideally is a significant error. However, this error is primarily localized in the pulmonary vein region that has inconsistent initial segmentation. The left atrium body, whose shape is indicative of AF recurrence, exhibits near sub-voxel accuracy. 

\textbf{Downstream Task - AF Recurrence Prediction: } As the focal loss does not make much difference, this downstream analysis is performed using the base- and TL-DeepSSM trained without the focal loss. Left atrium shape is indicative of AF recurrence \cite{harris2013cam}. In our dataset, we are also provided with binary outcome data that indicate whether or not the patient had AF recurrence post-ablation. Using this binary information,  we aim to train a supervised predictor to obtain AF recurrence probability from the shape descriptor. State-of-the-art PDM models provide us with correspondences, and an ideal choice of shape descriptor is to use PCA scores. However, not all the modes of PCA are useful in exhibiting anatomical changes that discriminate left atrium with and without AF recurrence. Hence, we perform feature trimming via a heuristic; we compute the absolute difference between the shape descriptor of the normal and AF recurrence left atrium and select ten dimensions with the highest magnitude. Using this trimmed PCA scores shape descriptor, we use a multi-layer perceptron for classification with 122 training and 84 testing data. We perform this for PCA scores from \emph{ShapeWorks} \cite{cates2017shapeworks} and Base-DeepSSM.
Furthermore, we also perform the same analysis utilizing the latent space of TL-DeepSSM as the shape descriptor. Both these models used are the ones without the focal loss. The results are showcased in Table \ref{tab:afacc}. We see that PDM and Base-DeepSSM perform similarly, indicating that the shape model from DeepSSM can be directly used for providing a useful shape-descriptor that performs comparably with the state-of-the-art PDM. We also see that TL-DeepSSM space outperforms the PCA scores space of other models; this can be attributed to the non-linear nature of the anatomical variation that is better captured by an autoencoder in place of PCA. Hence, the TL-DeepSSM learns a richer shape descriptor for the left atrium.

\begin{table}[]
\centering
\begin{tabular}{|l|l|l|l|}
\hline
\textbf{Model}         & \textbf{ShapeWorks}  & \textbf{Base-DeepSSM} & \textbf{TL-DeepSSM} \\ \hline
\textbf{Accuracy} & 61.9\%     & 59.5\% &    71.4\%     \\ \hline
\end{tabular}
\caption{\textbf{Accuracy of AF recurrence prediction: }The classification accuracy of AF recurrence utilizing the shape descriptors obtained from different models, ShapeWorks \cite{cates2017shapeworks}, Base- and TL-DeepSSM}
\label{tab:afacc}
\end{table}

\begin{figure}
    \centering
    \includegraphics[width=\textwidth]{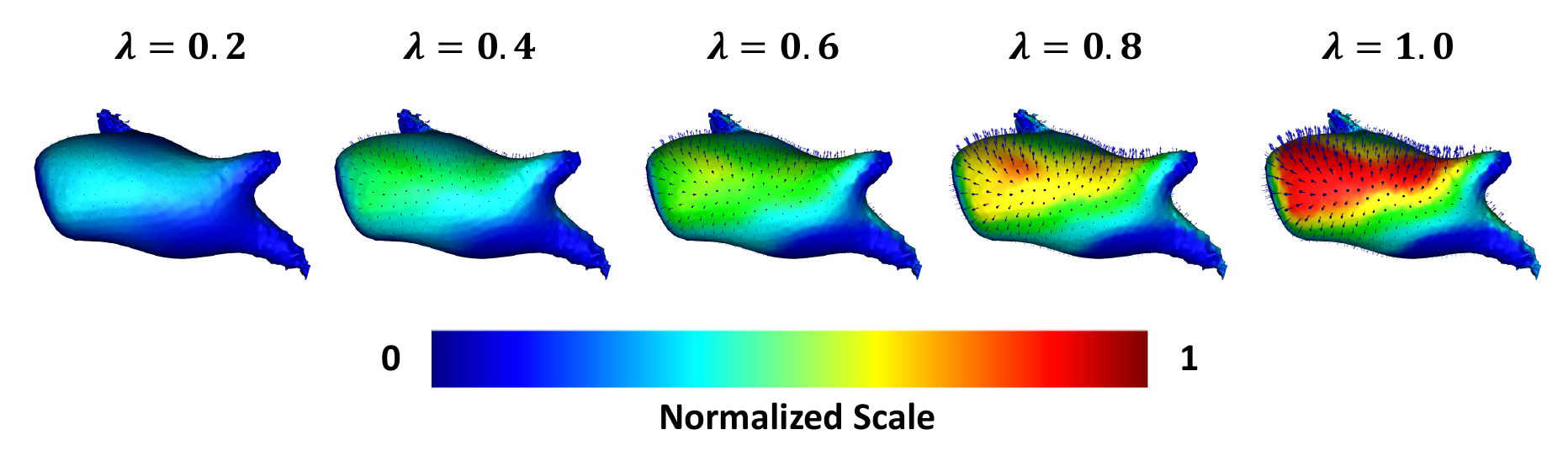}
    \caption{\textbf{TL-DeepSSM latent swim: } Samples taken from latent space of TL-DeepSSM trained on left atrium data, between the mean of normal shapes and mean of shape that exhibit AF recurrence after ablation. The difference between each sample and the mean normal is overlayed over the mean normal left atrium.}
    \label{fig:lda_la}
\end{figure}

Since TL-DeepSSM produces a more discriminatory shape descriptor for AF recurrence prediction, we move along the latent space of the TL-DeepSSM similar to analysis performed in Section \ref{sec:cam} in Figure \ref{fig:lda_femur}. Reiterating, we find the mean of the normal left atrium and mean of left atrium diagnosed with AF recurrence in the latent space. We move along the line (using parameter $\lambda \in [0, 1]$) joining these means and reconstruct the samples back to the correspondence space ($C_\lambda$). We take the difference of each of these samples from the normal mean and project the variation on the normal mean. Note that here we visualize $C_\lambda - C_0$ on $C_0$ shown in Figure \ref{fig:lda_la}, which is reverse of the visualization performed in Figure \ref{fig:lda_femur}, this is done for ease of visualization. Both ways still demonstrate the shape variation from normal to pathological shape. From Figure \ref{fig:lda_la} we see that as the $\lambda$ increases, i.e., moving towards pathology, the left atrium shows outward movement, indicating increased bulging of the left atrium as a morphological change observed in AF recurrence patients. This observation clearly shows the left atrium shape is more spherical in cases diagnosed with AF recurrence, this aligns with observations in previous studies \cite{bisbal2013left, bisbal2014reversal} that showcases that sphericity of the left atrium body is associated with recurrence, and the feature can be used to predict AF recurrence after ablation. Another work \cite{varela2017novel} observes a larger asymmetry in left atrium shapes of patients with AF recurrence, this also aligns with our observations in Figure \ref{fig:lda_la} that exhibits more expansion on one side than the other. This alignment with previous works and their observations strengthens the argument that TL-DeepSSM's latent space captures the anatomical variations that differentiate the left atrium shapes with and without AF recurrence after ablation, and can be effectively used as a predictor for the same.

\section{Discussion}
\label{sec:discussion}

The goal of DeepSSM is to arrive at a shape descriptor that adequately captures the shape variability in a population so that it can be used for downstream application without the need for resource-heavy processing. In this manuscript, via four different datasets and applications, we showcase that the correspondence model provided by DeepSSM and its variants is comparable in performance with state-of-the-art PDM, if not better in some scenarios. It validates the efficacy of DeepSSM and provides a novel way to achieve shape modeling tasks on new images without the need for segmentation, pre-processing, and optimization. Additionally, DeepSSM inference computationally surpasses the current methods by obtaining correspondences on a new scan in seconds as compared to ~1-2hrs when using \emph{ShapeWorks}\cite{cates2017shapeworks} and other PDMs. This processing-free and fast inference also allows shape model based applications to have a fully end-to-end pipeline that could be easily deployed, for instance, on cloud services. DeepSSM is not a single network; it is a customizable framework/methodology to get statistical shape models directly from images; it can be customized in architectural variant and loss functions. This manuscript aims to provide a comprehensive description of linear and non-linear architectures and related loss functions of DeepSSM and how they can be adapted for individual datasets/applications. 

\subsection{When to use each variant?} 
We have proposed two different loss functions as well as two different architectures. Throughout our experimentation, we have discovered that the utility of these variants differs based on the data/application the model is being used for. This discussion sheds light on where each variant might be helpful over others. 

As mentioned above, the choice for variant depends on the data or applications or both. Let us start by looking at applications of the shape descriptor. These can be broadly divided into two parts, (i) where the shape descriptor is being used as a feature for subsequent prediction/classification/clustering model, and (ii) where the feature is examined to understand the morphology of a sample/group of samples. In (i), we can use any shape descriptor, and it is on the subsequent model to extract the shape information. In case of (ii), the interpretability of the shape descriptor plays an important role, and PCA scores provide this, and hence, base DeepSSM will work better for such applications.  

Now let us look at the data, and here is where most of the subjectivity comes into play. For most datasets, TL-DeepSSM outperforms the base model (Figure \ref{fig:rmse_plots} and Figure \ref{fig:boxplts}). For datasets with limited variability in shapes and no natural clustering of morphology, the TL-DeepSSM performs marginally better quantitatively. This marginal difference does not affect the downstream application, and hence, it is better to use base DeepSSM, which provides us with a more interpretable shape descriptor. Whereas with anatomy like left atrium where we have highly non-linear shape variations, we see that the quantitative performance of TL-DeepSSM outshines other variants significantly. Furthermore, TL-DeepSSM's shape descriptor performs better at downstream tasks, showcasing that TL-DeepSSM should be used with such datasets. 
In some other datasets, the variation across the population is localized in some small regions and the rest are mostly constant. Such a scenario occurs with femur dataset (Section \ref{sec:cam}). In such cases utilizing focal loss, break the network out of the local-minima of learning data mean or few cluster means.

\section{Acknowledgments}

This work was supported by the National Institutes of Health under grant numbers NIBIB-U24EB029011, NIAMS-R01AR076120, NHLBI-R01HL135568, NIBIB-R21EB026061, NIBIB-R01EB016701, and NIGMS-P41GM103545. 
  
The content is solely the responsibility of the authors and does not necessarily represent the official views of the National Institutes of Health.

The authors would like to thank the Division of Cardiovascular Medicine (data were collected under Nassir Marrouche, MD, oversight and currently managed by Brent Wilson, MD, PhD) and the Orthopaedic Research Laboratory (ORL) (data were collected under Andrew Anderson, PhD, oversight) at the University of Utah for providing the MRI/CT scans and the corresponding segmentation of the left atrium and femur datasets, with a special thanks to Evgueni Kholmovski for assisting in MRI image acquisition. 

\appendix
\section{Network and Implementation Details}
\label{app:netdet}

\begin{figure}[!h]
    \centering
    \includegraphics[width=\textwidth]{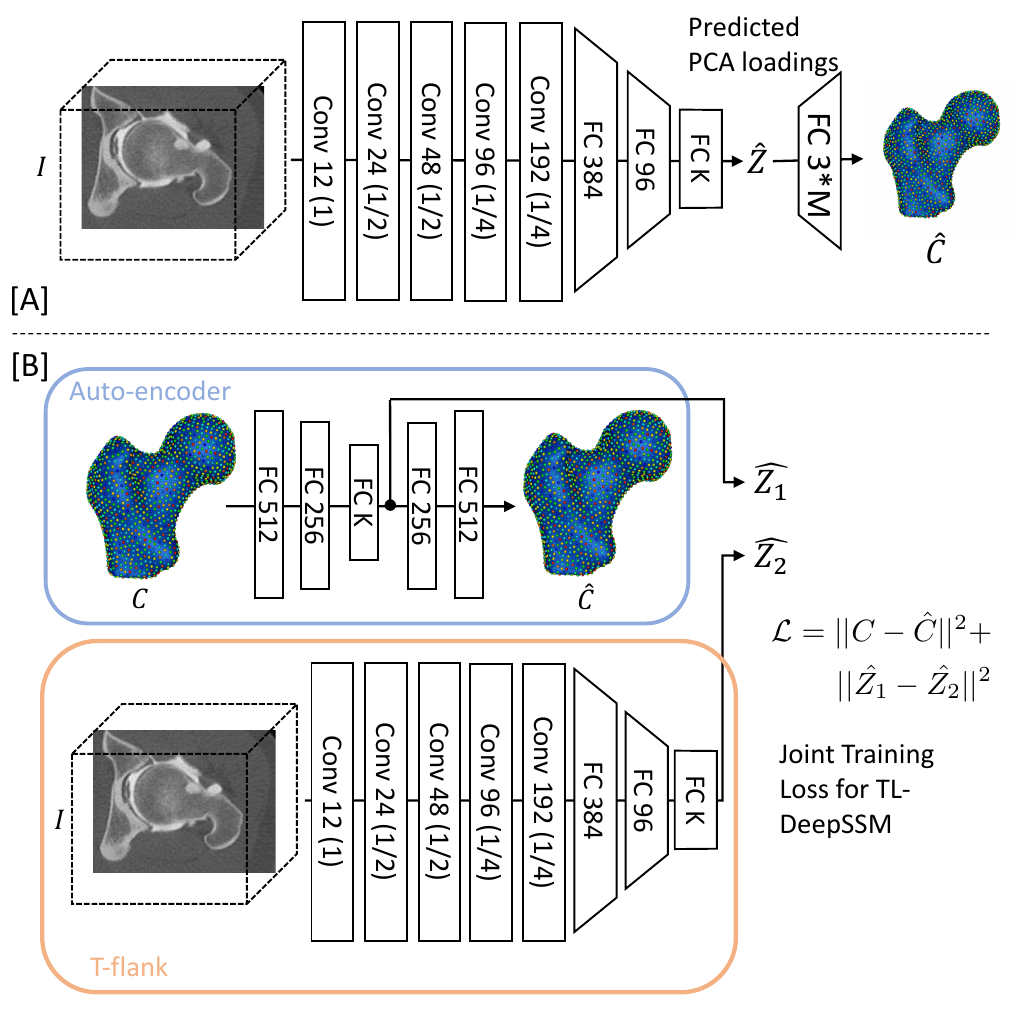}
    \caption{\textbf{Detailed network architectures for networks}}
    \label{fig:detarch}
\end{figure}

\subsection{Network Architecture}

The architecture for the base DeepSSM the last fully connected layer that performs the PCA reconstruction from the predicted loadings is kept fixed.
The architecture for the base model is shown in Figure \ref{fig:detarch}[A]. The latent encoder has five convolution layers with the number of output channels as [12, 24, 48, 96, 192], as shown in the figure, followed by three fully connected layers with output sizes of 384,96 and 
 K. Here, K is the number of PCA loadings being used for a given model. We have isotropic max-pooling by a factor of 2 after the first, third, and fifth convolution layers; this is denoted by the size factor shown at each block as (1), (1/2), and (1/4).  The output of this latent encoder is the predicted PCA loading shape descriptor. It is passed through a linear, fully connected layer, whose weights are initialized by the PCA basis and the bias by the mean shape. Again the parameters of this layer are \emph{fixed} for base DeepSSM; hence, it performs PCA reconstruction giving us the predicted correspondences. We use parametric ReLU \cite{he2015rectifiers} for as our non-linear activation function, and each layer is followed by batch normalization \cite{sergey2015BatchNormalization}. 

For the TL-DeepSSM (shown in Figure \ref{fig:detarch}[B]), the T-flank network is the same as the latent encoder for the base DeepSSM, and instead of producing the predicted PCA loadings, it will predict a K dimensional latent vector. The second part of the TL-DeepSSM is an autoencoder that consists of two fully connected layers for encoder and decoder each as shown in the figure, and the non-linear activation used here is leaky-ReLUs.

\subsection{Implementation Details}

The weights of the network as initialized by Xavier initialization \cite{xavier2010initialization}. The implementation of the networks was done in Pytorch, and experiments were performed on NVIDIA- TITAN V GPU with 12GB memory. In all cases, the batch size was selected to be between 5-10, and our experiments found that results are not sensitive to the batch size. Adam \cite{kingma2014adam} is used for optimization in all cases. For base DeepSSM experiments, we use an initial learning rate of 0.001 with a cosine annealing learning rate scheduler, along with a weight decay of 0.00005. For TL-DeepSSM, the autoencoder training was performed with a fixed learning rate of 0.0001 and a weight decay of 0.00005, followed by T-flank and joint training using an initial learning rate of 0.001 with cosine annealing and weight decay of 0.00005. Finally, we use early stopping for convergence by detecting the lowest validation loss found in the max number of epochs. All these optimization parameters are same for both the base particle loss Eq. \ref{eq:corr-l2} and for focal loss Eq \ref{eq:focal-l2}.

\bibliographystyle{splncs04}
\bibliography{references}

\begin{thebibliography}{10}
\providecommand{\url}[1]{\texttt{#1}}
\providecommand{\urlprefix}{URL }
\providecommand{\doi}[1]{https://doi.org/#1}

\bibitem{badrinarayanan2015segnet2}
Badrinarayanan, V., Handa, A., Cipolla, R.: Segnet: A deep convolutional
  encoder-decoder architecture for robust semantic pixel-wise labelling. arXiv
  preprint arXiv:1505.07293  (2015)

\bibitem{bajcsy1989multiresolution}
Bajcsy, R., Kova{\v{c}}i{\v{c}}, S.: Multiresolution elastic matching. Computer
  vision, graphics, and image processing  \textbf{46}(1),  1--21 (1989)

\bibitem{balakrishnan2018unsupervised}
Balakrishnan, G., Zhao, A., Sabuncu, M.R., Guttag, J., Dalca, A.V.: An
  unsupervised learning model for deformable medical image registration. In:
  CVPR. pp. 9252--9260 (2018)

\bibitem{beg2005computing}
Beg, M.F., Miller, M.I., Trouv{\'e}, A., Younes, L.: Computing large
  deformation metric mappings via geodesic flows of diffeomorphisms.
  International journal of computer vision  \textbf{61}(2),  139--157 (2005)

\bibitem{bhalodia2020quantifying}
Bhalodia, R., Dvoracek, L.A., Ayyash, A.M., Kavan, L., Whitaker, R., Goldstein,
  J.A.: Quantifying the severity of metopic craniosynostosis: a pilot study
  application of machine learning in craniofacial surgery. The Journal of
  craniofacial surgery  \textbf{31}(3), ~697 (2020)

\bibitem{bhalodia2018deepssm}
Bhalodia, R., Elhabian, S.Y., Kavan, L., Whitaker, R.T.: Deepssm: {A} deep
  learning framework for statistical shape modeling from raw images. In: Shape
  In Medical Imaging at MICCAI. Lecture Notes in Computer Science, vol. 11167,
  pp. 244--257. Springer (2018)

\bibitem{bhalodia2019cooperative}
Bhalodia, R., Elhabian, S.Y., Kavan, L., Whitaker, R.T.: A cooperative
  autoencoder for population-based regularization of cnn image registration.
  In: International Conference on Medical Image Computing and Computer-Assisted
  Intervention. pp. 391--400. Springer (2019)

\bibitem{bhalodia2018endtoend}
Bhalodia, R., Goparaju, A., Sodergren, T., Whitaker, R.T., Morris, A.,
  Kholmovski, E., Marrouche, N., Cates, J., Elhabian, S.Y.: Deep learning for
  end-to-end atrial fibrillation recurrence estimation. In: Computing in
  Cardiology, CinC 2018, Maastricht, The Netherlands, September 23-26, 2018
  (2018)

\bibitem{bieging2018left}
Bieging, E.T., Morris, A., Wilson, B.D., McGann, C.J., Marrouche, N.F., Cates,
  J.: Left atrial shape predicts recurrence after atrial fibrillation catheter
  ablation. Journal of cardiovascular electrophysiology  (2018)

\bibitem{bisbal2014reversal}
Bisbal, F., Guiu, E., Cabanas, P., Calvo, N., Berruezo, A., Tolosana, J.M.,
  Arbelo, E., Vidal, B., de~Caralt, T.M., Sitges, M., et~al.: Reversal of
  spherical remodelling of the left atrium after pulmonary vein isolation:
  incidence and predictors. Europace  \textbf{16}(6),  840--847 (2014)

\bibitem{bisbal2013left}
Bisbal, F., Guiu, E., Calvo, N., Marin, D., Berruezo, A., Arbelo, E.,
  Ortiz-P{\'e}rez, J., De~Caralt, T.M., Tolosana, J.M., Borr{\`a}s, R., et~al.:
  Left atrial sphericity: a new method to assess atrial remodeling. impact on
  the outcome of atrial fibrillation ablation. Journal of cardiovascular
  electrophysiology  \textbf{24}(7),  752--759 (2013)

\bibitem{cao2018deformable}
Cao, X., Yang, J., Zhang, J., Wang, Q., Yap, P.T., Shen, D.: Deformable image
  registration using a cue-aware deep regression network. IEEE Transactions on
  Biomedical Engineering  \textbf{65}(9),  1900--1911 (2018)

\bibitem{cates2017shapeworks}
Cates, J., Elhabian, S., Whitaker, R.: Shapeworks: Particle-based shape
  correspondence and visualization software. In: Statistical Shape and
  Deformation Analysis, pp. 257--298. Elsevier (2017)

\bibitem{cates2007shape}
Cates, J., Fletcher, P.T., Styner, M., Shenton, M., Whitaker, R.: Shape
  modeling and analysis with entropy-based particle systems. In: IPMI. pp.
  333--345. Springer (2007)

\bibitem{dalca2018anatomical}
Dalca, A.V., Guttag, J., Sabuncu, M.R.: Anatomical priors in convolutional
  networks for unsupervised biomedical segmentation. In: Proceedings of the
  IEEE Conference on Computer Vision and Pattern Recognition. pp. 9290--9299
  (2018)

\bibitem{davies2002MDL}
Davies, R.H., Twining, C.J., Cootes, T.F., Waterton, J.C., Taylor, C.J.: A
  minimum description length approach to statistical shape modeling. IEEE
  Transactions on Medical Imaging  \textbf{21}(5),  525--537 (May 2002).
  \doi{10.1109/TMI.2002.1009388}

\bibitem{durrleman2014morphometry}
Durrleman, S., Prastawa, M., Charon, N., Korenberg, J.R., Joshi, S., Gerig, G.,
  Trouv{\'e}, A.: Morphometry of anatomical shape complexes with dense
  deformations and sparse parameters. NeuroImage  \textbf{101},  35--49 (2014)

\bibitem{cates2013afib}
Gardner, G., Morris, A., Higuchi, K., MacLeod, R., Cates, J.: A
  point-correspondence approach to describing the distribution of image
  features on anatomical surfaces, with application to atrial fibrillation. In:
  2013 IEEE 10th International Symposium on Biomedical Imaging. pp. 226--229
  (April 2013). \doi{10.1109/ISBI.2013.6556453}

\bibitem{greig2001brain}
Gerig, G., Styner, M., Jones, D., Weinberger, D., Lieberman, J.: Shape analysis
  of brain ventricles using spharm. In: Proceedings IEEE Workshop on
  Mathematical Methods in Biomedical Image Analysis (MMBIA 2001). pp. 171--178
  (2001). \doi{10.1109/MMBIA.2001.991731}

\bibitem{girdhar2016TLnet}
Girdhar, R., Fouhey, D.F., Rodriguez, M., Gupta, A.: Learning a predictable and
  generative vector representation for objects. CoRR  \textbf{abs/1603.08637}
  (2016)

\bibitem{xavier2010initialization}
Glorot, X., Bengio, Y.: Understanding the difficulty of training deep
  feedforward neural networks. In: Proceedings of the Thirteenth International
  Conference on Artificial Intelligence and Statistics. Proceedings of Machine
  Learning Research, vol.~9, pp. 249--256. PMLR (13--15 May 2010)

\bibitem{goparaju2020benchmarking}
Goparaju, A., Bone, A., Hu, N., Henninger, H.B., Anderson, A.E., Durrleman, S.,
  Jacxsens, M., Morris, A., Csecs, I., Marrouche, N., et~al.: Benchmarking
  off-the-shelf statistical shape modeling tools in clinical applications.
  arXiv preprint arXiv:2009.02878  (2020)

\bibitem{goparaju2018evaluation}
Goparaju, A., Csecs, I., Morris, A., Kholmovski, E., Marrouche, N., Whitaker,
  R., Elhabian, S.: On the evaluation and validation of off-the-shelf
  statistical shape modeling tools: a clinical application. In: International
  Workshop on Shape in Medical Imaging. pp. 14--27. Springer (2018)

\bibitem{RTW:Gre91}
Grenander, U., Chow, Y., Keenan, D.M.: Hands: {A} Pattern Theoretic Study of
  Biological Shapes. Springer, New York (1991)

\bibitem{gutierrez2018deep}
Guti{\'e}rrez-Becker, B., Gatidis, S., Gutmann, D., Peters, A., Schlett, C.,
  Bamberg, F., Wachinger, C.: Deep shape analysis on abdominal organs for
  diabetes prediction. In: International Workshop on Shape in Medical Imaging.
  pp. 223--231. Springer (2018)

\bibitem{harris2013cam}
Harris, M.D., Datar, M., Whitaker, R.T., Jurrus, E.R., Peters, C.L., Anderson,
  A.E.: Statistical shape modeling of cam femoroacetabular impingement. Journal
  of Orthopaedic Research  \textbf{31}(10),  1620--1626 (2013).
  \doi{10.1002/jor.22389}, \url{http://dx.doi.org/10.1002/jor.22389}

\bibitem{he2015rectifiers}
He, K., Zhang, X., Ren, S., Sun, J.: Delving deep into rectifiers: Surpassing
  human-level performance on imagenet classification. CoRR
  \textbf{abs/1502.01852} (2015), \url{http://arxiv.org/abs/1502.01852}

\bibitem{ho2012left}
Ho, S.Y., Cabrera, J.A., Sanchez-Quintana, D.: Left atrial anatomy revisited.
  Circulation: Arrhythmia and Electrophysiology  \textbf{5}(1),  220--228
  (2012)

\bibitem{huang2017heartnet}
Huang, W., Bridge, C.P., Noble, J.A., Zisserman, A.: Temporal heartnet: Towards
  human-level automatic analysis of fetal cardiac screening video. In: MICCAI
  2017. pp. 341--349. Springer International Publishing (2017)

\bibitem{sergey2015BatchNormalization}
Ioffe, S., Szegedy, C.: Batch normalization: Accelerating deep network training
  by reducing internal covariate shift. In: Proceedings of the 32Nd
  International Conference on International Conference on Machine Learning -
  Volume 37. pp. 448--456. ICML'15, JMLR.org (2015),
  \url{http://dl.acm.org/citation.cfm?id=3045118.3045167}

\bibitem{joshi2012diffeomorphic}
Joshi, S.H., Cabeen, R.P., Joshi, A.A., Sun, B., Dinov, I., Narr, K.L., Toga,
  A.W., Woods, R.P.: Diffeomorphic sulcal shape analysis on the cortex. IEEE
  transactions on medical imaging  \textbf{31}(6),  1195--1212 (2012)

\bibitem{kellogg2012interfrontal}
Kellogg, R., Allori, A.C., Rogers, G.F., Marcus, J.R.: Interfrontal angle for
  characterization of trigonocephaly: part 1: development and validation of a
  tool for diagnosis of metopic synostosis. Journal of Craniofacial Surgery
  \textbf{23}(3),  799--804 (2012)

\bibitem{kingma2014adam}
Kingma, D.P., Ba, J.: Adam: A method for stochastic optimization (2017)

\bibitem{kozic2010optimisation}
Kozic, N., Weber, S., B{\"u}chler, P., Lutz, C., Reimers, N., Ballester,
  M.{\'A}.G., Reyes, M.: Optimisation of orthopaedic implant design using
  statistical shape space analysis based on level sets. Medical image analysis
  \textbf{14}(3),  265--275 (2010)

\bibitem{lecun1998cnn}
Lecun, Y., Bottou, L., Bengio, Y., Haffner, P.: Gradient-based learning applied
  to document recognition. Proceedings of the IEEE  \textbf{86}(11),
  2278--2324 (Nov 1998). \doi{10.1109/5.726791}

\bibitem{li2014classification}
Li, Q., Cai, W., Wang, X., Zhou, Y., Feng, D.D., Chen, M.: Medical image
  classification with convolutional neural network. In: 2014 13th International
  Conference on Control Automation Robotics Vision (ICARCV). pp. 844--848 (Dec
  2014). \doi{10.1109/ICARCV.2014.7064414}

\bibitem{lin2017focal}
Lin, T.Y., Goyal, P., Girshick, R., He, K., Doll{\'a}r, P.: Focal loss for
  dense object detection. In: Proceedings of the IEEE international conference
  on computer vision. pp. 2980--2988 (2017)

\bibitem{lu2018deep}
Lu, X., Ma, C., Ni, B., Yang, X., Reid, I., Yang, M.H.: Deep regression
  tracking with shrinkage loss. In: Proceedings of the European conference on
  computer vision (ECCV). pp. 353--369 (2018)

\bibitem{milletari2017stats}
Milletari, F., Rothberg, A., Jia, J., Sofka, M.: Integrating statistical prior
  knowledge into convolutional neural networks. In: Descoteaux, M.,
  Maier-Hein, L., Franz, A., Jannin, P., Collins, D.L., Duchesne, S. (eds.)
  Medical Image Computing and Computer Assisted Intervention. pp. 161--168.
  Springer International Publishing, Cham (2017)

\bibitem{muthen1998statistical}
Muth{\'e}n, L.K., Muth{\'e}n, B.O.: Statistical analysis with latent variables.
  Mplus User’s guide  \textbf{2012} (1998)

\bibitem{ACNN}
Oktay, O., Ferrante, E., Kamnitsas, K., Heinrich, M.P., Bai, W., Caballero, J.,
  Guerrero, R., Cook, S.A., de~Marvao, A., Dawes, T., O'Regan, D.P., Kainz, B.,
  Glocker, B., Rueckert, D.: Anatomically constrained neural networks {(ACNN):}
  application to cardiac image enhancement and segmentation. CoRR
  \textbf{abs/1705.08302} (2017), \url{http://arxiv.org/abs/1705.08302}

\bibitem{ovsjanikov2012functional}
Ovsjanikov, M., Ben-Chen, M., Solomon, J., Butscher, A., Guibas, L.: Functional
  maps: a flexible representation of maps between shapes. ACM Transactions on
  Graphics (TOG)  \textbf{31}(4), ~30 (2012)

\bibitem{ronneberger2015unet}
Ronneberger, O., Fischer, P., Brox, T.: U-net: Convolutional networks for
  biomedical image segmentation. CoRR  \textbf{abs/1505.04597} (2015),
  \url{http://arxiv.org/abs/1505.04597}

\bibitem{rueckert1999nonrigid}
Rueckert, D., Sonoda, L.I., Hayes, C., Hill, D.L.G., Leach, M.O., Hawkes, D.J.:
  Nonrigid registration using free-form deformations: application to breast mr
  images. IEEE Transactions on Medical Imaging  \textbf{18}(8),  712--721
  (1999)

\bibitem{RTW:Sty2000}
Styner, M., Brechbuhler, C., Szekely, G., Gerig, G.: Parametric estimate of
  intensity inhomogeneities applied to {MRI}. IEEE Transactions on Medical
  Imaging  \textbf{19}(3),  153--165 (Mar 2000)

\bibitem{styner2006spharm}
Styner, M., Oguz, I., Xu, S., Brechbuehler, C., Pantazis, D., Levitt, J.,
  Shenton, M., Gerig, G.: Framework for the statistical shape analysis of brain
  structures using spharm-pdm  (07 2006)

\bibitem{cates2018afib}
T., B.E., Alan, M., D., W.B., J., M.C., F., M.N., Joshua, C.: Left atrial shape
  predicts recurrence after atrial fibrillation catheter ablation. Journal of
  Cardiovascular Electrophysiology  \textbf{0}(0). \doi{10.1111/jce.13641}

\bibitem{tao2020unsupervised}
Tao, W., Bhalodia, R., Anstadt, E., Kavan, L., Whitaker, R.T., Goldstein, J.A.:
  Unsupervised shape normality metric for severity quantification. arXiv
  preprint arXiv:2007.09307  (2020)

\bibitem{thompson1942growth}
Thompson, D.W., et~al.: On growth and form. On growth and form.  (1942)

\bibitem{Tothova2020}
T{\'{o}}thov{\'{a}}, K., Parisot, S., Lee, M.C.H., Puyol{-}Ant{\'{o}}n, E.,
  King, A.P., Pollefeys, M., Konukoglu, E.: Probabilistic 3d surface
  reconstruction from sparse {MRI} information. CoRR  \textbf{abs/2010.02041}
  (2020), \url{https://arxiv.org/abs/2010.02041}

\bibitem{Tothova2018}
T{\'{o}}thov{\'{a}}, K., Parisot, S., Lee, M.C.H., Puyol{-}Ant{\'{o}}n, E.,
  Koch, L.M., King, A.P., Konukoglu, E., Pollefeys, M.: Uncertainty
  quantification in cnn-based surface prediction using shape priors. CoRR
  \textbf{abs/1807.11272} (2018), \url{http://arxiv.org/abs/1807.11272}

\bibitem{uebersax1993latent}
Uebersax, J.S., Grove, W.M.: A latent trait finite mixture model for the
  analysis of rating agreement. Biometrics pp. 823--835 (1993)

\bibitem{varela2017novel}
Varela, M., Bisbal, F., Zacur, E., Berruezo, A., Aslanidi, O.V., Mont, L.,
  Lamata, P.: Novel computational analysis of left atrial anatomy improves
  prediction of atrial fibrillation recurrence after ablation. Frontiers in
  physiology  \textbf{8}, ~68 (2017)

\bibitem{wang2019deeply}
Wang, B., Lei, Y., Tian, S., Wang, T., Liu, Y., Patel, P., Jani, A.B., Mao, H.,
  Curran, W.J., Liu, T., et~al.: Deeply supervised 3d fully convolutional
  networks with group dilated convolution for automatic mri prostate
  segmentation. Medical physics  \textbf{46}(4),  1707--1718 (2019)

\bibitem{yang2017quicksilver}
Yang, X., Kwitt, R., Styner, M., Niethammer, M.: Quicksilver: Fast predictive
  image registration--a deep learning approach. NeuroImage  \textbf{158},
  378--396 (2017)

\bibitem{zhao2008hippocampus}
Zhao, Z., Taylor, W.D., Styner, M., Steffens, D.C., Krishnan, K.R.R., MacFall,
  J.R.: Hippocampus shape analysis and late-life depression. PLoS One
  \textbf{3}(3),  e1837 (2008)

\bibitem{zheng2015detection}
Zheng, Y., Liu, D., Georgescu, B., Nguyen, H., Comaniciu, D.: 3d deep learning
  for efficient and robust landmark detection in volumetric data. In: MICCAI
  2015. pp. 565--572. Springer International Publishing (2015)

\end{thebibliography}

\end{document}